\pdfoutput=1
\documentclass[runningheads]{llncs}

\usepackage{eccv}

\newcommand{\tablestyle}[2]{\setlength{\tabcolsep}{#1}\renewcommand{\arraystretch}{#2}\centering\footnotesize}

\usepackage{eccvabbrv}

\usepackage{graphicx}
\usepackage{booktabs}

\usepackage[accsupp]{axessibility}

\usepackage{hyperref}

\usepackage{orcidlink}

\usepackage{wrapfig}
\usepackage{xcolor}
\usepackage{colortbl}
\usepackage{multirow}
\usepackage{multicol}
\usepackage{colortbl}

\definecolor{colFull}{HTML}{D6EAF8}
\definecolor{colLoRA}{HTML}{D9F2D0}
\definecolor{colLoKr}{HTML}{FFFFD5}
\definecolor{colLoHa}{HTML}{CAEEFB}
\definecolor{colDelta}{HTML}{EAF0FB}

\begin{document}

\title{Dive into the implicit biases of low-rank vision-language alignment
}

\titlerunning{Dive into the implicit biases of low-rank vision-language alignment
}

\author{Mingjia Shi\orcidlink{0000-0002-9988-3741}
Shuo Wang\inst{1}$^{\ddag}$\orcidlink{0009-0003-9907-6186}
Xiaobo Wang\inst{2}$^{\dag}$
Sifan Zhou\inst{1}\orcidlink{0000-0003-3602-7566}
Kai Wang\inst{3}\orcidlink{0000-0002-1154-5175} \\
Tianyu Fu\inst{1}\orcidlink{0009-0001-7656-350X}
Chenxu Zhao\inst{1}$^{\dag}$\orcidlink{0000-0003-4044-5701}
Anyang Su\inst{1}\orcidlink{0000-0002-0085-3527}
Ping Jiang\inst{1}\orcidlink{0009-0008-6161-060X}
Minghui Wu\inst{1}\orcidlink{0009-0001-4577-3002}
}

\authorrunning{M.~Shi et al.}

\institute{Mininglamp \and
Shenzhen University of Advanced Technology
\and
National University of Singapore
\\
\email{\{wangshuo.e,zhaochenxu\}@mininglamp.com, 3101ihs@gmail.com, wangxiaobo@suat-sz.edu.cn}}

\maketitle
\footnotetext{$^{\dag}$\,Corresponding author. \quad $^{\ddag}$\,Project lead.}

\begin{abstract}
Vision-language alignment, the stage that bridges pretrained vision encoders and large language models, is widely treated as a form of pretraining requiring full-parameter updates. We challenge this view and investigate what happens when low-rank adaptation is applied to the LLM during this stage instead. We find that low-rank alignment not only reduces computational costs but also outperforms full-parameter alignment on most benchmarks. To understand this phenomenon, we systematically characterize the implicit biases introduced by low-rank adaptation during alignment. Empirically, we find that low-rank alignment shifts model behavior from hallucinatory to conservative and preserves per-token linear separability of visual features that full-parameter alignment disrupts, a phenomenon we term \emph{LS-curse}. Geometrically, low-rank aligned models exhibit more homogeneous and structurally stable visual representations, maintaining modality-specific knowledge rather than prematurely fusing entity-level semantics. Theoretically, we establish two theorems showing that low-rank alignment induces preferences for parameter subspaces with flat gradients and feature subspaces robust to perturbations, providing a principled explanation for the observed structure-preserving behavior. Extensive experiments cover ablation over 100 alignment configurations, three families of low-rank operators, and various rank, encoder, and other settings.

\end{abstract}

\section{Introduction}
\label{sec:intro}
\begin{figure}[t]
    \centering
    \includegraphics[width=.98\linewidth]{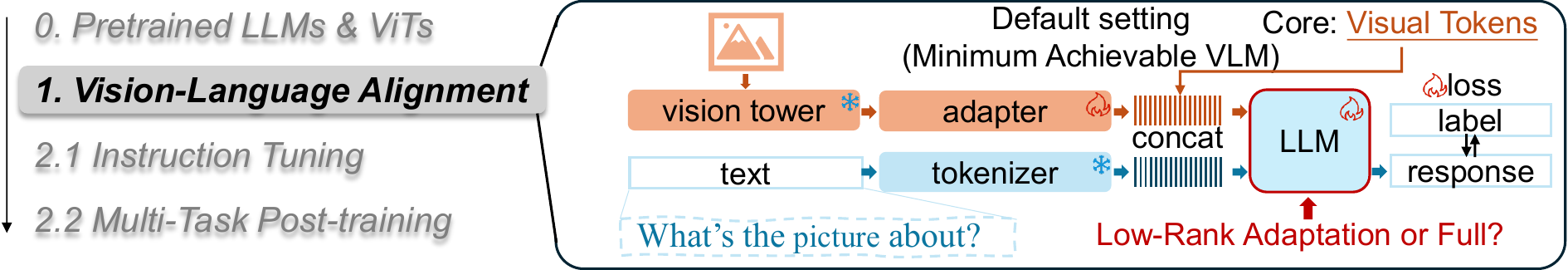}
    \vspace{-10pt}
    \caption{Vision-language alignment framework (default): a minimum achievable VLM with a cross-modal adapter (MLP). How does \textit{low-rank alignment} shape \textit{visual tokens}?.}
    \vspace{-20pt}
    \label{fig:framework}
\end{figure}

The construction of vision-language models (VLMs) hinges on a critical stage: \emph{vision-language alignment}, where pretrained vision encoders are bridged with large language models (LLMs) through adapter modules and joint training~\cite{liu2023improvedllava,chen2024sharegpt4v,chu2024mobilevlm2}. As illustrated in Figure~\ref{fig:framework}, this stage precedes instruction tuning and establishes the cross-modal feature correspondence upon which all downstream capabilities depend. Standard practice updates the full LLM parameters during alignment, treating it as a form of pretraining. However, this view constitutes a misconception: unlike LLM pretraining, which builds linguistic representations from scratch, vision-language alignment operates atop pretrained knowledge and reasoning priors: it is, in fact, \emph{supervised fine-tuning}, precisely the regime for which low-rank adaptation methods were designed~\cite{hulora,han2024parameter,zhang2023instruction,dong2026visnec}. Yet, applying low-rank methods during this stage remains largely unexplored.

When low-rank adaptation is applied to the LLM during alignment (\textit{i.e.}, \textit{low-rank alignment}), we observe that not only does the training cost drop substantially, but downstream performance \emph{empirically improves} across all involved model scales (from 1.4B to 14B). This improvement holds across diverse benchmarks spanning perception, knowledge, reasoning, and hallucination, and generalizes across three families of low-rank operators (LoRA, LoHa, LoKr). This finding raises a natural question: \textbf{how do the low-rank methods reshape the visual features to yield superior representations, and technically, what \emph{implicit biases} does the \textit{low-rank alignment} introduce?}

\textbf{Empirical observations.}
We investigate this question first at the behavior and feature levels. At the behavior level, hypothesis testing on visual perception benchmarks reveals that low-rank alignment shifts model decision-making from hallucinatory to conservative, reducing overconfident predictions while preserving cautious uncertainty (\S\ref{sec:behavioral}). At the feature level, token-wise linear probes uncover that full-parameter alignment disrupts the linear separability of certain visual tokens, a phenomenon we term \emph{LS-curse}, whereas low-rank alignment preserves the per-token knowledge structure inherited from the pretrained vision tower (\S\ref{sec:feature}). Together, these observations suggest that the performance gain stems from low-rank alignment's stronger preservation of modality-specific knowledge structures, rather than the premature entity-level mixing induced by full-parameter updates.

\textbf{Geometric characterization.}
We further characterize these biases through geometric analysis of the visual feature space. Fine-grained and coarse-grained linear probes reveal that fully aligned models exhibit greater feature diversity and broader angular coverage across tokens, reflecting aggressive entity-level fusion. In contrast, low-rank aligned models maintain homogeneous and structurally conservative representations. Manifold visualizations corroborate this finding: per-token feature distributions under low-rank alignment are markedly more consistent than those under full alignment (\S\ref{sec:geometric}).

\textbf{Theoretical analysis.}
We establish two theorems characterizing the mathematical mechanisms underlying these biases. Theorem~\ref{thm:noise-smoothing} reveals that the low-rank gradient flow exhibits a noise-weighted smoothing effect, preferentially reinforcing updates along feature directions robust to stochastic variations, encoding an implicit bias toward flatter regions of the loss landscape. Theorem~\ref{thm:steady-state} demonstrates that the steady-state distribution of low-rank parameters concentrates on subspaces with flat gradients and noise-robust features. These two results jointly explain why low-rank alignment preserves general knowledge structures instead of prematurely fusing entity-specific characteristics (\S\ref{sec:theory}).

We conduct extensive experiments across over 55 vision-language alignment settings (unique combinations of model, vision encoder, low-rank operator, rank, vision-tower depth, and learning-rate schedule) spanning model architectures from MobileLLaMA-1.4B to Qwen3-14B and three families of low-rank operators, which together yield over 100 configurations when evaluated across benchmark categories.
We summarize our contributions as follows:
\begin{itemize}
    \item We demonstrate that low-rank adaptation during vision-language alignment not only improves efficiency but also consistently enhances downstream performance across all evaluated model scales (1.4B–14B), challenging the prevailing assumption that alignment requires full-parameter updates.
    \item We provide a systematic empirical and geometric characterization of the implicit biases introduced by low-rank alignment, identifying the LS-curse in full alignment and revealing how low-rank methods preserve modality-specific knowledge structures, collectively supporting a \emph{progressive fusion} view in which general visual representations are retained during alignment and entity-specific fusion is deferred to instruction tuning.
    \item We establish two theorems identifying the theoretical mechanisms: noise-robust feature preferences during optimization (Theorem~\ref{thm:noise-smoothing}), and flat-gradient subspace concentration at steady state (Theorem~\ref{thm:steady-state}), providing a principled explanation for the structure-preserving behavior observed empirically.
    \item We conduct comprehensive ablations covering rank, operators, model scales, architectures, encoder generalization (CLIP, DINOv2, SigLIPv2) and other details, providing practical guidance for efficient multimodal training.
\end{itemize}

\section{Empirical Study: Low-Rank Alignment's Preferences}
\label{sec:empirical}
\paragraph{Training protocol.}
\begin{wrapfigure}[12]{r}{0.5\textwidth}
    \centering
    \vspace{-30pt}
    \includegraphics[width=\linewidth]{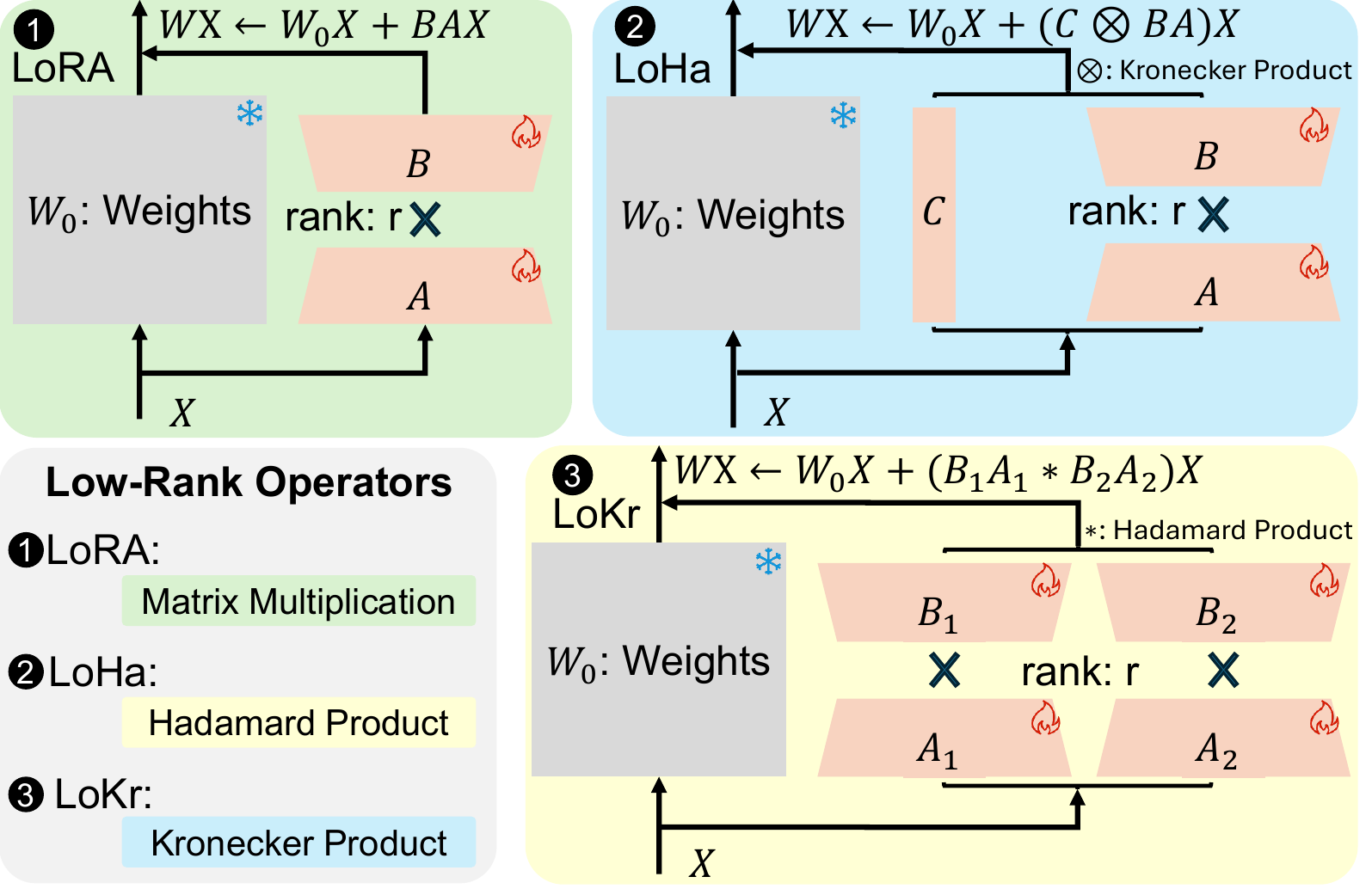}
    \vspace{-20pt}
    \caption{Operators for low-rank alignment: LoRA/LoHa/LoKr (Matrix, Hadamard and Kronecker Products).}
    \label{fig:low-rank-op}
    \vspace{-15pt}
\end{wrapfigure}

As illustrated in Figure~\ref{fig:framework}, during the vision-language alignment stage, we employ perceptual question answering (QA) tasks with general prompting templates. Trainable parameters comprise the LLM backbone and the vision-tower adapter; the vision encoder is frozen by default. For low-rank alignment, we apply adaptation exclusively to all linear modules in the LLM, since the adapters are already lightweight; imposing low-rank constraints on them would force all vision features into a low-rank subspace. Both full-parameter and low-rank baselines are tuned under matched hyperparameter budgets to ensure fair comparison (best out of 9 times). For behavioral evaluation, instruction tuning is necessary to ensure proper instruction following. To assess zero-shot capabilities, we divide instruction tuning into two phases: initial tuning with a base visual question answering (VQA) instruction dataset, followed by post-training with an extended instruction dataset covering task-specific and knowledge-intensive data. All benchmarks except GQA and TextVQA are treated as unseen tasks. Extended ablations on encoder choice, frozen \textit{vs.}~unfrozen settings, low-rank placement, learning rate schedules, and overfitting controls are provided in \S\ref{sec:ablation}.

\paragraph{Low-rank operators.}
We focus on linear transformations, specifically LoRA~\cite{hulora}, LoHa~\cite{loha}, and LoKr~\cite{lokr}, which encompass the three principal product-based constructions for low-rank modules: matrix multiplication, Hadamard product, and Kronecker product. The formulations and operator details are in Figure~\ref{fig:low-rank-op}.

\paragraph{Linear probe.}
The procedure extracts features from a classification dataset and trains a linear classifier on them to measure accuracy. This approach enables us to assess both the robustness of the corresponding knowledge and the degree of feature entanglement.

\paragraph{Resources.}
We conduct experiments across a range of language models spanning compact to commonly used scales, including MobileLLaMA-1.4/2.7B~\cite{chu2024mobilevlm2}, Vicuna-7/13B~\cite{chiang2023vicuna}, and Qwen3-8/14B\footnote{Detailed experiments regarding Qwen3 and other settings are in the Appendix.}~\cite{yang2025qwen3technicalreport}. The vision tower is CLIP-Vision~\cite{clip}. For vision-language alignment, we utilize datasets including LAION~\cite{schuhmann2022laion}, CC~\cite{sharma2018conceptual}, SBU~\cite{ordonez2011im2text}, SAM~\cite{ravi2024sam}, and MS-COCO~\cite{lin2014microsoft}. Instruction tuning leverages default instruction datasets, including LLaVA~\cite{liu2023improvedllava}, ShareGPT4V-PT~\cite{chen2024sharegpt4v}, GQA~\cite{hudson2019gqa}, MS-COCO-train-2017, OCRVQA~\cite{mishra2019ocr}, TextVQA~\cite{singh2019textvqa}, and VG~\cite{krishna2017visual}, along with extended instruction datasets: SBU, ScienceQA~\cite{SQA}, MS-COCO,
GQA, IconQA~\cite{lu2021iconqa}, SAM, ShareGPT4V-TextVQA~\cite{chen2024sharegpt4v}, Web-Celebrity~\cite{liu2015faceattributes}, Web-Landmark~\cite{chu2024mobilevlm2}, and WikiArt~\cite{wikiart2022}. Evaluation is performed on a diverse set of benchmarks and tasks, including GQA, MMBench~\cite{liu2024mmbench}, MME~\cite{yin2024survey}, MMMU-Pro-Vision~\cite{yue2025mmmu}, MMVet-gpt~\cite{yu2024mm}, POPE~\cite{POPE}, ScienceQA, and TextVQA.

\paragraph{Default settings.}
Since this work focuses on the alignment stage, we minimize the VLM pipeline to its essential components, deliberately removing unnecessary architectural tricks to facilitate clean ablation and isolate alignment-stage effects. Unless otherwise specified, instruction fine-tuning uses GQA, TextVQA, and OCRVQA as task-specific datasets, and LoRA with rank 128 by default. For clarity in reporting, we categorize the main benchmarks into four groups: \emph{perception},~\emph{knowledge},~\emph{reasoning},~\text{and}~\emph{hallucination}, corresponding to MME-perception, GQA, MME-reasoning, and POPE, respectively. For simplicity, we refer to models by their size rather than their full names. Default visualizations and fine-grained analyses are performed using the 1.4B model to reduce the computational and storage overhead associated with complex processing (\textit{e.g.}, feature extraction and dimensionality reduction).

\subsection{Pre-Experiments: Performance of Low-Rank Alignment}\label{sec:pre-exp}

\begin{table*}[t]
\centering
\tablestyle{0.1pt}{1.3}
  \caption{Performance of low-rank vs.\ full-parameter alignment across model scales (1.4B--13B). Low-rank results report the best configuration across LoRA, LoHa, and LoKr after three rounds of rank tuning. Metrics cover perception (MME-P), knowledge (GQA), reasoning (MME-R), and hallucination (POPE). Training time on 8$\times$A100.}
  \label{tbl:pre-exp2}
  \vspace{-10pt}
  \begin{tabular}{@{}rrrrrrrrrrr@{}}
    \cellcolor{gray!20}{size}
        &
         \emph{perception}$\uparrow$ &\cellcolor{colDelta}{$\Delta$ $\uparrow$}
         &
         \emph{knwldg.}$\uparrow$ &\cellcolor{colDelta}{$\Delta$ $\uparrow$}
         &
         \emph{reason}$\uparrow$ &\cellcolor{colDelta}{$\Delta$ $\uparrow$}
         &
         \emph{hallu.}$\uparrow$ &\cellcolor{colDelta}{$\Delta$ $\uparrow$}
         &
         time (h) $\downarrow$
         &
         \cellcolor{colDelta}{$-\Delta$ $\uparrow$}
         \\
        \hline
         \cellcolor{gray!20}{1.4B}
         &208.1[\underline{453.5}] &\cellcolor{colDelta} 245.4 &18.9[\underline{28.0}] &\cellcolor{colDelta} 9.1 &22.9[\underline{766.6}] &\cellcolor{colDelta} 743.7 &2.5[\underline{72.8}] &\cellcolor{colDelta} 70.3 &4.53[\underline{2.52}] &\cellcolor{colDelta} 2.01\\

         \cellcolor{gray!20}{2.7B}
         &15.7[\underline{669.0}] &\cellcolor{colDelta} 653.3 &11.6[\underline{42.4}]& \cellcolor{colDelta} 30.8 &40.4[\underline{201.4}] &\cellcolor{colDelta} 161.0 &0.5[\underline{76.4}] &\cellcolor{colDelta} 75.9 & 8.61[\underline{4.29}] &\cellcolor{colDelta} 4.32\\

         \cellcolor{gray!20}{7B}
         & 521.0[\underline{965.3}]& \cellcolor{colDelta} 444.3& 23.4[\underline{38.5}]& \cellcolor{colDelta} 15.1&86.8[\underline{332.5}] &\cellcolor{colDelta} 245.7 & 35.0[\underline{80.4}] & \cellcolor{colDelta} 45.4&19.75[\underline{7.68}] & \cellcolor{colDelta} 12.07 \\

         \cellcolor{gray!20}{13B}
         & 236.6[\underline{1079.7}] &\cellcolor{colDelta} 843.1 & 29.5[\underline{43.0}] & \cellcolor{colDelta} 13.5 & 52.5[\underline{219.6}] & \cellcolor{colDelta} 167.1 & 66.9[\underline{82.9}] & \cellcolor{colDelta} 16.0 &35.32[\underline{13.23}] & \cellcolor{colDelta} 21.09 \\
  \end{tabular}
  \vspace{-20pt}
\end{table*}

We evaluate models aligned via low-rank methods under the default instruction-tuning setting described above. Table~\ref{tbl:pre-exp2} demonstrates that low-rank vision-language alignment substantially enhances final model performance across all scales, even when implemented solely through low-rank branches prior to instruction tuning. It raises the question: \textit{why and how does low-rank alignment boost performance?} We then address this question through both experimental and theoretical analyzes of the mechanism of low-rank alignment.

\subsection{Behavioral Preferences: From Hallucinatory to Conservative}\label{sec:behavioral}

We first examine whether low-rank alignment alters model decision-making behavior. To this end, we evaluate aligned models on a yes-or-no visual QA dataset and measure Type~I and Type~II error rates after instruction tuning. In this context, a Type~I error occurs when the model incorrectly answers ``yes'' to a negative-ground-truth question, indicating an overconfident or hallucinated response. A Type~II error occurs when the model incorrectly answers ``no'' to a positive-ground-truth question, reflecting cautious behavior that defaults to rejection under uncertainty. At the sentence level, such cautiousness manifests as increased perplexity in the generated answers.

\begin{wrapfigure}[11]{r}{0.6\textwidth}
    \centering
    \vspace{-25pt}
    \includegraphics[width=\linewidth]{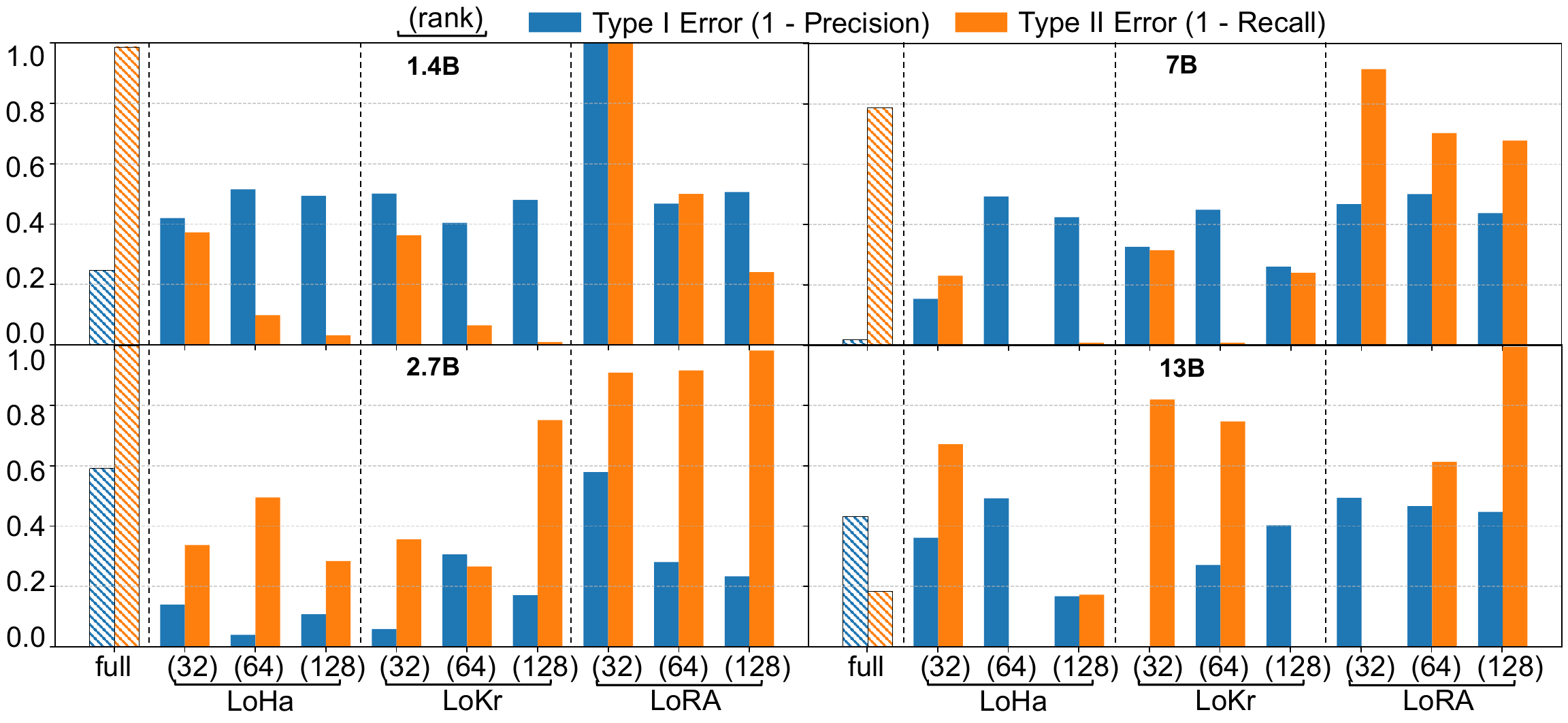}
    \vspace{-20pt}
    \caption{Type -I and -II error rate: models and low-rank operators. The results are on POPE, a visual perception hallucination benchmark.}
    \vspace{-10pt}
    \label{fig:error_rates}
\end{wrapfigure}

Figure~\ref{fig:error_rates} demonstrates that low-rank adaptation shifts model behavior from hallucinatory to conservative. At the 1.4B and 7B scales, LoHa substantially reduces the Type~I error rate compared to full fine-tuning, effectively reversing the model's behavioral tendency from overconfident affirmation toward cautious rejection. Through grid search over rank values, we find that certain low-rank operators, such as LoHa, can simultaneously reduce both Type~I and Type~II error rates without inducing a mere trade-off between the two failure modes. We use \emph{conservative} to denote a reduced Type~I (hallucination) rate and \emph{decisive} to denote that this reduction is not obtained at the cost of a higher Type~II rate, \textit{i.e.}, both error modes decrease jointly rather than trading off. We note that this shift is context-dependent: at the smallest (1.4B) scale the gap narrows, since an under-capacity model already tends toward conservative default decisions, and conservatism is not universally beneficial: settings that require affirmative identification may favor less cautious behavior.
\begin{center}
\vspace{-5pt}
\boxed{Obs. \hypertarget{obs-1}{1}~\text{Low-rank alignment is associated with conservative/decisive behaviors.}}
\vspace{-5pt}
\end{center}

\subsection{Feature Preferences: Linear Separability Preservation}\label{sec:feature}
We next investigate how the visual features (\textit{i.e.}, the LLM's visual inputs) are shaped by alignment before multimodal fusion takes place. We employ ImageNet~\cite{russakovsky2015imagenetlargescalevisual} as the test dataset to evaluate general vision capabilities.

\begin{wrapfigure}[10]{t}{0.5\linewidth}
    \centering
    \vspace{-20pt}
\includegraphics[width=\linewidth]{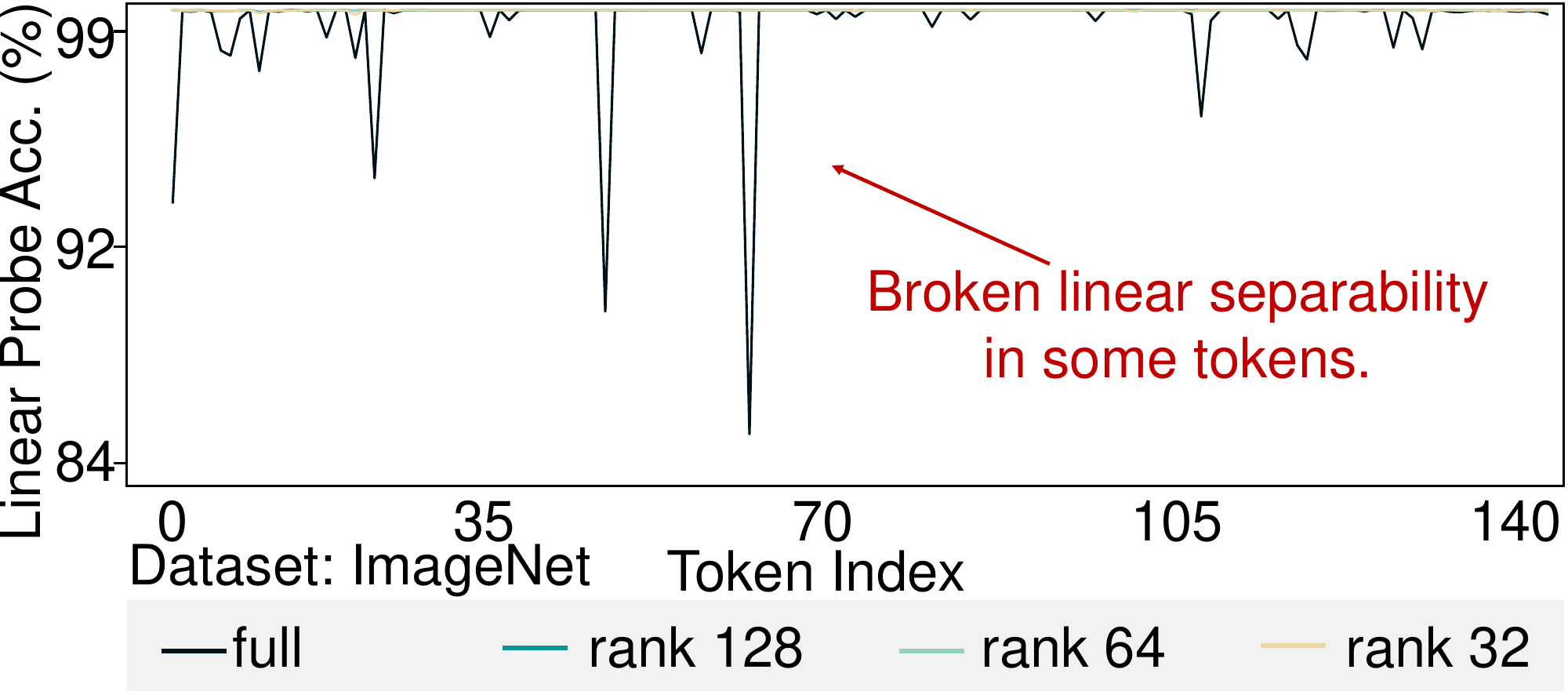}
    \vspace{-20pt}
    \caption{Broken per-token linear separability of visual tokens, marked in Figure~\ref{fig:framework}, tested by the default settings on ImageNet.}
    \label{fig:lora_linear_probe}
\end{wrapfigure}

We measure \emph{per-token linear separability} (LS), which assesses whether each visual token independently retains a linearly separable knowledge structure. Concretely, for each token position $t$ in the visual sequence, we train a separate linear classifier on the corresponding $C$-dimensional feature vectors across all samples and report classification accuracy. A high per-token LS indicates that the knowledge encoded at that position remains well-structured and discriminative; a low value suggests that the token's representation has been entangled with information from other tokens or modalities.

The original vision tower is well-trained and exhibits near-perfect token-wise linear separability. Figure~\ref{fig:lora_linear_probe} reveals that in a low-level perception setting (\textit{i.e.}, a 1.4B model with full or low-rank alignment using a 2-layer vision-language connector), the vision tokens of aligned models before instruction tuning no longer maintain full linear separability on ImageNet. For the low-rank aligned model (LoRA with rank 128), individual tokens remain largely linearly separable, whereas full-parameter alignment disrupts this property in a subset of tokens, a phenomenon we term \emph{LS-curse}: certain tokens lose their discriminative structure entirely, with accuracy dropping from near-perfect to as low as 83.6\%. As shown in the following table, while some fully aligned tokens retain the knowledge structure from generic image datasets, others exhibit a mixing of representations that destroys linear separability.
\begin{table}[h]
\tablestyle{8pt}{1.3}
    \centering
    \vspace{-20pt}\begin{tabular}{cccc}
    &
    origin
    &
    full
    &
    low-rank
    \\
    \hline
    avg. acc. $\pm$ std. (\%)~[\textcolor{red!70}{min}]
    &
    100$\pm$0.00
    &
    \textcolor{blue!60}{99.52}$\pm${1.89}~[\textcolor{red!70}{83.61}]
    &
    \textcolor{blue!60}{99.99}$\pm$0.02
    \vspace{-20pt}
    \end{tabular}
\end{table}

This mixing of partial representations is not necessarily detrimental; it can introduce beneficial diversity across token representations. For instance, adversarial samples with similar representations may align semantically, thus appearing non-adversarial in certain tasks or contexts. However, it reflects a fundamentally different alignment strategy: full-parameter alignment aggressively restructures the visual feature space, whereas low-rank alignment operates conservatively within a constrained subspace.
\begin{center}
\vspace{-5pt}
\boxed{Obs. \hypertarget{obs-2}{2}~\text{Low-rank alignment preserves per-token linear separability.}}
\vspace{-5pt}
\end{center}

\paragraph{Interpretation.}
The combination of Obs.~\hyperlink{obs-1}{1} and \hyperlink{obs-2}{2} reveals a coherent picture of the implicit biases introduced by low-rank alignment. At the behavior level, low-rank aligned models exhibit greater conservatism and reduced susceptibility to hallucination. At the feature level, this conservatism is grounded in the preservation of modality-specific knowledge structures: low-rank alignment maintains the per-token linear separability inherited from the pretrained vision tower, whereas full alignment disrupts this structure through aggressive entity-level mixing. The subspace-constrained nature of low-rank updates limits token-wise diversity, preserving commonalities within each modality and preventing premature fusion of entity-level characteristics, while adjusting the subspace that most strongly affects the majority of features to maximize alignment loss reduction.

\begin{table}[ht]
\vspace{-5pt}
\tablestyle{7pt}{1.3}
    \centering
    \vspace{-20pt}
    \begin{tabular}{lcc}
         &
         full
         &
         low-rank
         \\
         \hline
         (behavior) decisiveness
         &
         hallucinatory
         &
         conservative
         \\
         (feature) preference
         &
         entity characteristics
         &
         modal commonality
    \end{tabular}
    \vspace{-30pt}\label{tab:empirical_hypothesis_summary}
\end{table}

\subsection{Geometric Characterization}\label{sec:geometric}

To further characterize the structural differences between full and low-rank alignment, we conduct two token-level geometric analyses: (1) testing each token's linear separability on diverse, unseen fine-grained classification datasets (\textit{e.g.}, birds and dogs) to disentangle general from entity-specific knowledge; and (2) examining the angular coverage and manifold structure of token features on general coarse-grained datasets (\textit{e.g.}, ImageNet).

\begin{table*}[t]
\caption{Linear probe results across datasets of varying knowledge granularity. The fully aligned (full) vision tower's features are more diverse (wider range of linear probe acc. on fine-grained datasets). ImageNet is general classification dataset. STF-dogs and CUB are fine-grained classification dataset of dogs and birds. Ranks of LoRA are 32/64/128. Range is reported by the maximum and minimum.}
\tablestyle{6.5pt}{1.3}
  \label{tab:linear-probe-specific-knowledge}
  \centering
  \vspace{-10pt}
  \begin{tabular}{@{}lcrrrrr@{}}
   dataset
        &
        fine-grained
        &
        full
        &
        rank=128
        &
        rank=64
        &
        rank=32
        \\
        \hline
        ImageNet
        &
        \texttimes
        &
        83.6$\sim$100.0
        &
        99.9$\sim$100.0
        &
        99.8$\sim$100.0
        &
        99.8$\sim$100.0
        \\
        STF-dogs
        &
        \checkmark
        &
        32.8$\sim$95.7
        &
        39.2$\sim$88.0
        &
        36.3$\sim$85.9
        &
        36.7$\sim$86.8
        \\
        CUB
        &
        \checkmark
        &
        49.9$\sim$94.0
        &
        61.2$\sim$91.7
        &
        60.8$\sim$91.5
        &
        61.6$\sim$91.5
         \\
  \end{tabular}
  \vspace{-20pt}
\end{table*}

\paragraph{General \textit{vs.}~specific knowledge.}
If low-rank alignment preserves general knowledge structures without prematurely fusing entity-specific characteristics, the two alignment types should yield comparable per-token LS on fine-grained entity categories while differing substantially on coarse-grained, general-purpose datasets. Table~\ref{tab:linear-probe-specific-knowledge} confirms this prediction. On both STF-dogs~\cite{khosla2011novel} and CUB~\cite{wah2011caltech} (fine-grained datasets targeting dogs and birds, respectively), the fully aligned model exhibits high peak accuracy (\textit{e.g.}, 95.7\% on STF-dogs, 94.0\% on CUB) but also pronounced inter-token variability, with the worst tokens degrading significantly. The low-rank aligned model, by contrast, maintains narrower and more uniform LS distributions: it trades peak fine-grained separability for consistent preservation of coarse, foundational representations across all tokens.
\begin{wrapfigure}[15]{r}{0.6\textwidth}
    \centering
    \includegraphics[width=\linewidth]{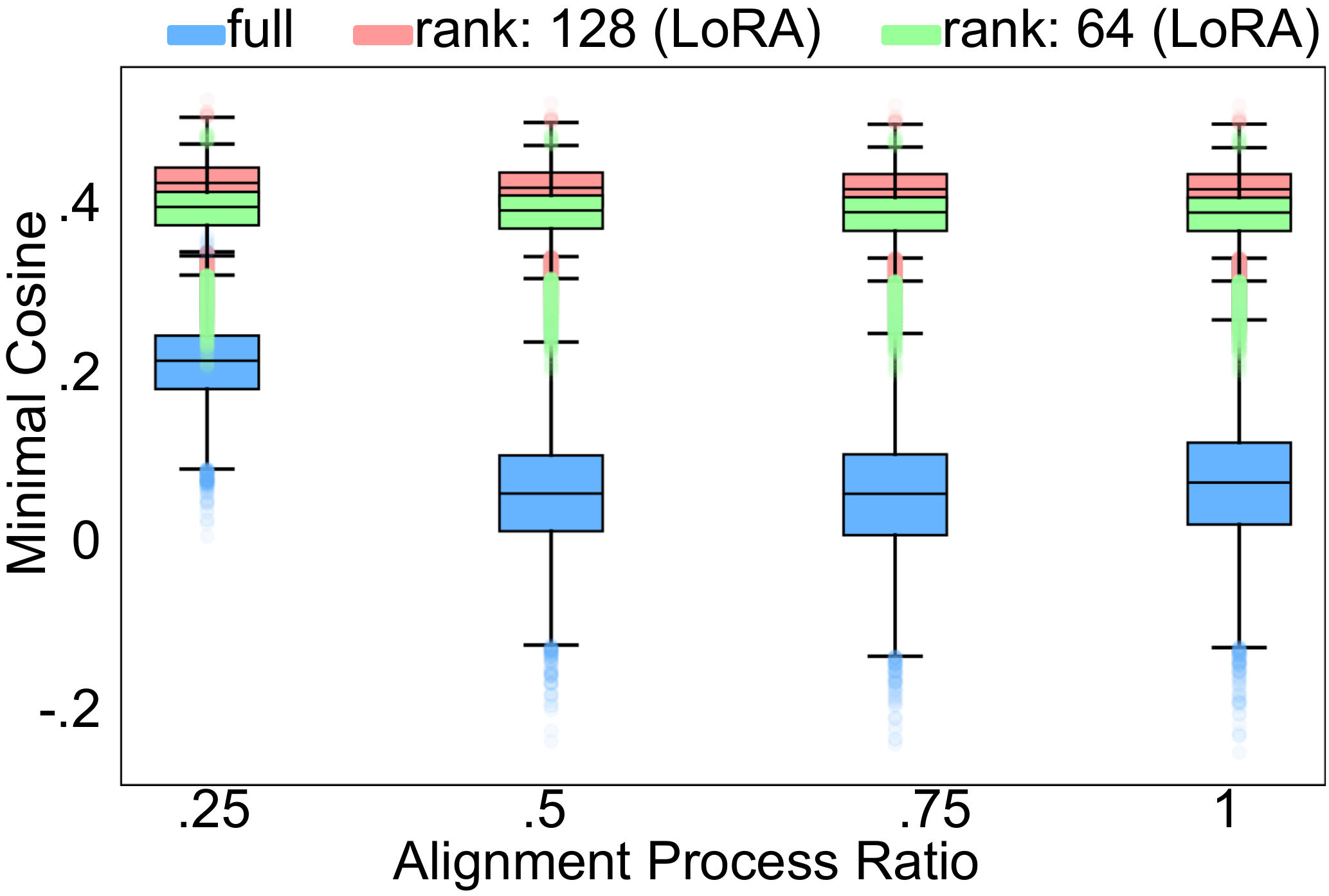}
    \vspace{-20pt}
    \caption{Full aligned vision tokens have larger coverage of each given input images in ImageNet. Each image's feature angle coverage (\textit{i.e.}, minimal cosine $\downarrow$ :=$\min_{i,j}\cos(i,j),i,j\in\{\mathrm{visual tokens}|\mathrm{images}\}$).}
    \vspace{-30pt}
    \label{fig:cos-process}
\end{wrapfigure}

This asymmetry directly supports the progressive fusion hypothesis: low-rank alignment retains the general knowledge backbone while deferring entity-specific fusion to subsequent tuning stages.

\paragraph{Representation geometry.}
We quantify how alignment reshapes the angular structure of the visual feature space. For each entity sampled from ImageNet, we compute the minimum pairwise cosine angle among its corresponding visual tokens; a smaller angle indicates that tokens have spread farther apart on the hyperspherical surface. As shown in Figure~\ref{fig:cos-process}, the horizontal axis tracks alignment progress and the vertical axis reports this angular coverage.
The fully aligned model covers a progressively wider angular span, at times exceeding half of the hyperspherical surface, indicating aggressive spread and fusion of token representations. The low-rank aligned model, by contrast, maintains a compact angular footprint throughout, preserving the original structural layout of the visual features.
\vspace{-2mm}

\begin{wrapfigure}[15]{r}{0.6\textwidth}
    \centering
    \includegraphics[width=\linewidth]{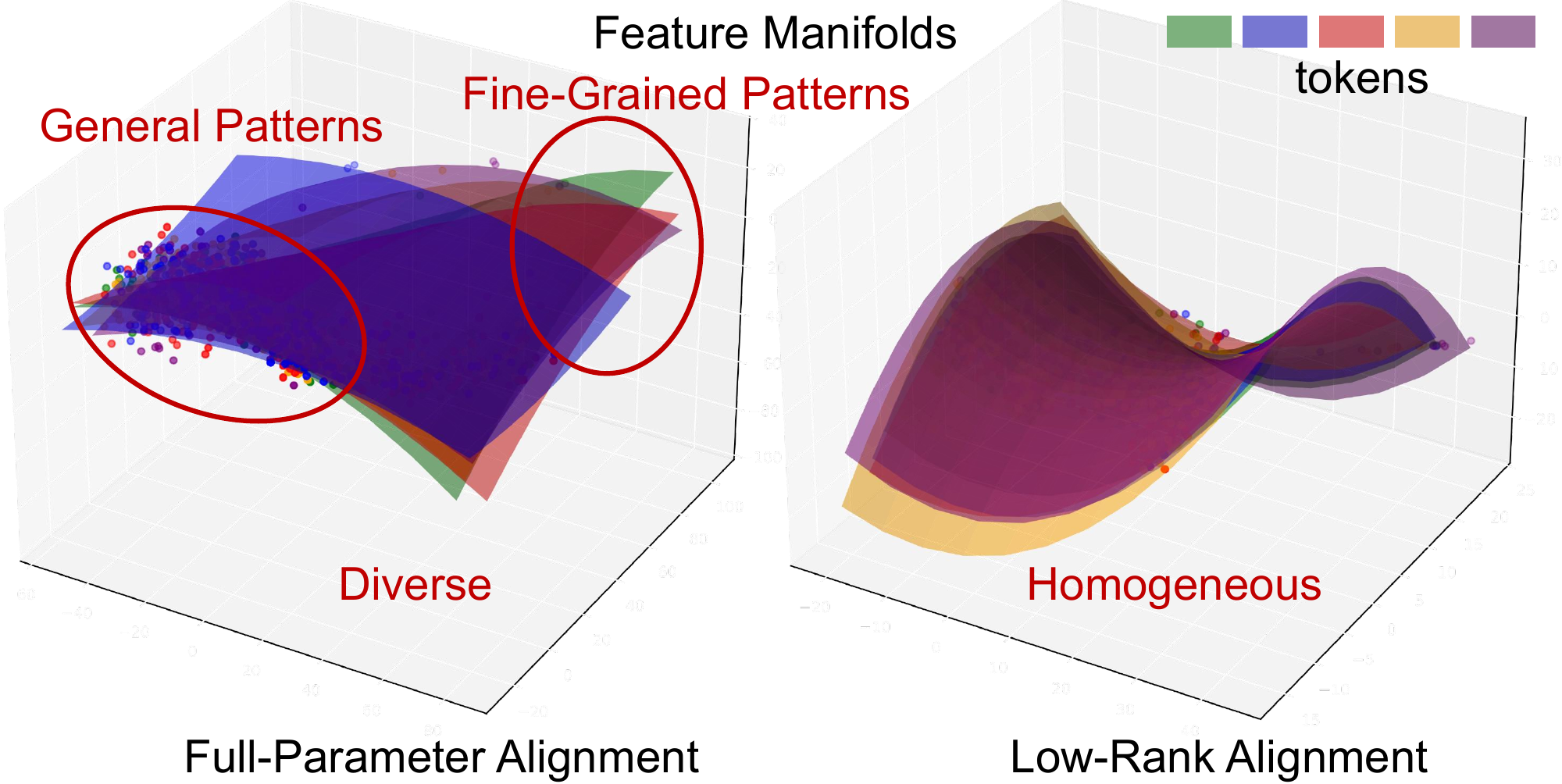}
    \vspace{-20pt}
    \caption{The feature manifolds of each visual token, as depicted in Figure~\ref{fig:framework}, are diverse under full alignment and homogeneous under low-rank alignment (LoRA, rank 128). The surface is fitted using a polynomial, retaining first- and second-order information. Each colored manifold represents the token feature distribution on MS-COCO.}
    \label{fig:manifold}
\end{wrapfigure}

\paragraph{Feature manifold.} Figure~\ref{fig:manifold} visualizes the token-wise feature manifolds (where colors represent different tokens), corroborating the angular analysis above. Under full alignment, per-token manifolds are geometrically diverse and spatially dispersed across the feature space, reflecting aggressive restructuring of visual representations. Under low-rank alignment, token clusters are markedly more homogeneous in shape and collectively occupy a more compact region, confirming that the low-rank constraint induces a structural bias toward conservative, stable representations. This geometric regularity is consistent with updates confined to a dominant low-rank subspace, preserving the spatial organization inherited from the pre-trained encoder rather than aggressively overwriting it.

\paragraph{Empirical summary.}
Low-rank alignment introduces systematic implicit biases that manifest at multiple levels. At the behavior level, it produces conservative and decisive perception. At the feature level, it preserves per-token knowledge structure and prevents the LS-curse. Geometrically, it maintains homogeneous and structurally stable visual representations while preventing the premature spread and overgeneralization of entity-level features observed in full alignment. These converging observations motivate the theoretical analysis in Section~\ref{sec:theory}, where we identify the mathematical mechanisms underlying these biases.

\vspace{-3mm}
\section{Theoretical Study: Preferences on Parameter and Representation Spaces, Flatness and Robustness}
\label{sec:theory}
\vspace{-2mm}
The empirical study in Section~\ref{sec:empirical} reveals that low-rank alignment exhibits systematic implicit biases: conservative behavior, preservation of per-token knowledge structure, and geometrically stable visual representations. Theoretical analysis is provided to identify the sufficient conditions under which such patterns arise and derive their mathematical characterizations. \emph{Due to the presence of higher-order tensors beyond second order in our derivations, we adopt the Einstein summation convention by default, wherein duplicated indices imply summation.}

\paragraph{Analysis model: unconstrained features.} We adopt the Unconstrained Feature Model~\cite{mixon2022neural} (UFM) as our analytical framework. UFM assumes that an over-parameterized neural network can fit any feature, thereby treating features $H$ as free variables rather than fixed outputs of a specific architecture. This enables us to study global properties of the learned feature structure, such as which directions are preferentially reinforced. In practice, this assumption is well-supported by the over-parameterized nature of modern LLMs used in the alignment stage.

\paragraph{Low-rank modeling.} Consider a low-rank adaptation parameterized by $\theta$
with forward mapping $O(\theta,H)$ and total loss
\vspace{-2mm}
\begin{align*}
F(\theta) &= \mathbb{E}_{(x,y)\sim \mathcal{D}} \left[ \mathcal{L}(O(\theta,H(x)),y) \right],
\\
O^i &= W_0{}^{i}{}_{j} \, H^{j} + b^{i} + l^{i}(H;\theta),
\end{align*}
where $O^i \in \mathbb{R}^p$ and $H^j \in \mathbb{R}^q$ denote the output and input features indexed by $i$ and $j$; $W_0^{\,i}{}_{j} \in \mathbb{R}^{p \times q}$ and $b^i \in \mathbb{R}^p$ are the frozen base weight and bias; $l^i(H;\theta)$ is the low-rank adaptation term; $\mathcal{D}$ is the training dataset; and $\mathcal{L}$ is the loss function. For LoRA, $l^{i}(H;\theta)=B^{i}_{k}A^{k}_{j}H^{j}$ with low-rank factors $A^{\,k}{}_{j} \in \mathbb{R}^{r \times q}$, $B^{\,i}{}_{k} \in \mathbb{R}^{p \times r}$, and $\theta = \{B,A\}$. Analogous parameterizations for LoHa and LoKr, involving Hadamard factor pairs $B_1, B_2 \in \mathbb{R}^{p \times r}$, $A_1, A_2 \in \mathbb{R}^{r \times q}$ and Kronecker factor $C^{\,s}{}_{t} \in \mathbb{R}^{s \times t}$, are detailed in the Appendix.

\subsection{Process Preference: Robustness to Feature Noise}\label{sec:theorem-robust}

\vspace{-3mm}
\begin{theorem}[Feature preferences: noise smoothing.]
\label{thm:noise-smoothing}
Under the UFM assumption, the feature space with additive noise exhibits a weighted smoothing effect.
Expanding features with zero-mean noise \(\xi\), we have
\vspace{-2mm}
\begin{align*}
\mathbb{E}_\xi\left[ \frac{\partial F}{\partial W_{0}^{i}{}_{j}} \right]
\simeq \frac{\partial F}{\partial W_{0}^{i}{}_{j}}(\bar H)
+ \frac{1}{2} \frac{\partial^3 F}{\partial W_{0}^{i}{}_{j} \, \partial H^k \, \partial H^l} \, \Sigma^{kl},
\end{align*}
where \(\Sigma^{kl} = \mathbb{E}[\xi^k \xi^l]\) is the feature noise covariance, $F$ is the alignment loss function, $W_{0}$ is the original weight, $\bar{H}$ the mean of $H:=\bar{H}+\xi$.
\end{theorem}
\begin{corollary}[LoRA's\footnote{Corollaries about LoHa and LoKr can be obtained. See Appendix.} noise robustness preference
.]
\label{corollary:robustness_preference}
    Substituting this into the LoRA gradient flow yields

\vspace{-2mm}
\begin{align*}
\dot{A}^{p}{}_{j} &\simeq -\eta B^{i}{}_{p}
\left(
\frac{\partial F}{\partial W_{0}^{i}{}_{j}}(\bar H)  + \frac{1}{2} \frac{\partial^3 F}{\partial W_{0}^{i}{}_{j} \, \partial H^k \, \partial H^l} \, \Sigma^{kl}
\right),
\\
\dot B^i{}_p
&\simeq -\eta \left(
\frac{\partial F}{\partial W_{0}^i{}_j}(\bar H)
+ \frac{1}{2} \frac{\partial^3 F}{\partial W_{0}^i{}_j \partial H^k \partial H^l} \Sigma^{kl}
\right) A^p{}_j,
\end{align*}
demonstrating an implicit smoothing effect along directions with feature noise.
This shows that gradient flow preferentially aligns updates along subspaces that are robust to stochastic variations in the features, encoding an implicit bias towards flatter directions in the loss landscape.
\end{corollary}
\vspace{-2mm}

\paragraph{Intuition and explanation.} The core insight lies in the coupled gradient flow: $\dot{A} \sim B \cdot (\text{grad}+\text{noise})$ and $\dot{B} \sim (\text{grad}+\text{noise}) \cdot A$, where $\dot{A}$ and $\dot{B}$ denote the gradient flows of $A$ and $B$, respectively, and $*$ and $\otimes$ denote the Hadamard and Kronecker products where applicable. Since updates to $A$ are scaled by $B$ and vice versa, whichever directions are already dominant receive proportionally larger updates, a self-reinforcing dynamic. The noise-induced third-order term further amplifies this effect along directions with large feature-noise covariance, creating a Matthew effect: strong directions grow stronger while weak directions are suppressed. Consequently, LoRA naturally concentrates updates on noise-robust principal directions in the feature space.

\paragraph{Comparison: full \textit{vs.}~low-rank alignment.} In full alignment, the weight matrix $W$ is updated directly and all directions receive gradient updates on an equal footing. In low-rank alignment, $A$ and $B$ act as mutual subspace projectors, confining the effect of third-order smoothing terms to the principal subspace. This confinement favors conservative feature integration along dominant directions while suppressing changes in weaker subspaces, helping to explain the homogeneous manifold structure and preserved linear separability observed in Section~\ref{sec:feature}. Here $\delta_i^j$ denotes the Kronecker delta arising in the projection identities. Analogous conclusions hold for LoHa and LoKr as detailed in the Appendix.

\subsection{Steady-State Preferences: Flatness and Robustness}\label{sec:theorem-flat}

Theorem~1 characterizes which directions are reinforced \emph{during} optimization. We now ask a complementary question: \emph{where} do the low-rank parameters converge at the end of training?

\begin{theorem}[Low-rank steady-state distribution]
\label{thm:steady-state}
Let the gradient flow of $\theta$ be
$\dot \theta^\alpha = -\eta \frac{\partial F}{\partial \theta^\alpha}$.
Define the effective low-rank subspace mapping $MH=l(H)$ (for example, $M = BA$ in LoRA, $M=B_{1}A_{1}*B_{2}A_{2}$ in LoHa, $M=C\otimes BA$ in LoKr) and the corresponding effective loss:
\vspace{-2mm}
\begin{align*}
F_{\rm eff}(M) = F(M)
+ \frac{\eta \log \det \widetilde{D}(M)}{4}
+ \frac{\partial^2 F}{2\partial O^i \partial O^m} \,
  \Sigma^{jk}(W_0^i{}_j + M^i{}_j)(W_0^m{}_k + M^m{}_k),
\end{align*}
where $F(M)$ is the original loss expressed in the low-rank subspace $M$; $\widetilde D(M)$ is the covariance of gradient noise projected onto $M$, encoding flatness preference; $\Sigma^{jk}$ is the covariance of feature noise. Then, under small-step stochastic gradient descent or noisy gradient flow, the steady-state $M$ distributes approximately as:
\vspace{-2mm}
\begin{align*}
p_{\rm stat}(M) \propto \exp\Big(-\frac{2}{\eta} F_{\rm eff}(M) \Big).
\end{align*}
\end{theorem}

\begin{remark}
The effective loss $F_{\rm eff}$ augments $F$ with two penalty terms. The log-determinant term penalizes subspaces where gradient noise is small or isotropic, biasing convergence toward flat regions of the loss landscape. The feature-noise term penalizes subspaces sensitive to input perturbations, biasing convergence toward noise-robust directions. Together, they ensure that steady-state parameters concentrate on subspaces that are simultaneously flat and robust.
\end{remark}

\paragraph{Intuition and takeaways.}
The two penalty terms reveal a dual implicit bias: low-rank parameters are drawn toward configurations that are flat with respect to the loss and robust with respect to input perturbations. The net effect is that the learned subspace naturally avoids directions sensitive to small variations: the system settles into configurations where outputs remain stable under feature noise, inherently reducing susceptibility to adversarial perturbations and aligning with the conservative behavior observed empirically in Section~\ref{sec:empirical}.

\section{Ablation Study: Low-rank Vision-Language Alignment}
\label{sec:ablation}
\paragraph{Overview.}
The ablation studies investigate the effects of rank, model scale, alignment depth, low-rank operators, vision encoder choice, frozen \textit{vs.}~unfrozen encoder settings, low-rank placement, learning rate schedules, and overfitting controls. Unless otherwise specified, experiments follow the default settings with targeted modifications. Extended results are provided in the Appendix.

\begin{table}[t]
\centering
\caption{Comparison of between low-rank operators on Vicuna-13B. Full-parameter alignment serves as the baseline. The selection of low rank operators is highly correlated with rank. A suitable range of rank can be empirically found with better perception.}
\vspace{-10pt}
\label{tab:ranks}
\tablestyle{5.4pt}{1.3}
\begin{tabular}{l
    >{\columncolor{colFull}}c
    >{\columncolor{colLoRA}}c >{\columncolor{colLoRA}}c >{\columncolor{colLoRA}}c
    >{\columncolor{colLoKr}}c >{\columncolor{colLoKr}}c >{\columncolor{colLoKr}}c
    >{\columncolor{colLoHa}}c >{\columncolor{colLoHa}}c >{\columncolor{colLoHa}}c
}
& full
& \multicolumn{3}{c}{\cellcolor{colLoRA}{LoRA}}
& \multicolumn{3}{c}{\cellcolor{colLoKr}{LoKr}}
& \multicolumn{3}{c}{\cellcolor{colLoHa}{LoHa}} \\
metric & \cellcolor{colFull}--
& \cellcolor{colLoRA}32 & \cellcolor{colLoRA}64 & \cellcolor{colLoRA}128
& \cellcolor{colLoKr}32 & \cellcolor{colLoKr}64 & \cellcolor{colLoKr}128
& \cellcolor{colLoHa}32 & \cellcolor{colLoHa}64 & \cellcolor{colLoHa}128 \\
\hline
MME-P & 236.6 & 375.9 & 288.0 &  -- &  79.4 &  78.5 & 449.3 & 380.6 & 621.4 & \textbf{1079.7} \\
POPE  &  42.2 &  50.0 &  19.7 &   -- &   9.0 &  12.7 &  46.8 &  21.6 &  51.5 &   \textbf{82.8} \\
\end{tabular}
\vspace{-10pt}
\end{table}

\paragraph{Ranks.}
In perception-intensive benchmarks, increasing the rank of the low-rank module yields significant performance gains, particularly for larger models. This is evident from the improvements on MME-perception and POPE using the 13B model
In Table~\ref{tab:ranks}
, in contrast, knowledge-intensive or task-specific benchmarks (\textit{e.g.}, GQA, ScienceQA, TextVQA) show no comparable gains from rank increases, suggesting that the performance bottleneck in these tasks lies elsewhere. We explore this further in the extended-data ablation in the Appendix.

\paragraph{Operators.}
\begin{table}[t]
    \centering
    \caption{Ablation on low-rank operators across model scales. Results are reported on 7B / 13B MobileVLM-V2 models. Best results in \textbf{bold}.}
    \vspace{-10pt}
    \tablestyle{5pt}{1.3}
    \begin{tabular}{l
        >{\columncolor{white}}c
        >{\columncolor{white}}c
        >{\columncolor{white}}c
        >{\columncolor{white}}c
    }
         operators on {7}/{13}B
         &
         \emph{perception$\uparrow$}
         &
         \emph{knowledge$\uparrow$}
         &
         \emph{reasoning$\uparrow$}
         &
         \emph{hallucination$\uparrow$}
         \\
         \hline
         \rowcolor{colFull}
         full parameter
         &
         {521.0}/{236.6}
         &
         {23.4}/{29.5}
         &
         {86.8}/{52.5}
         &
         {35.0}/{66.9}
         \\
         \rowcolor{colLoRA}
         LoRA: mat. mul.
         &
         {551.8}/{375.9}
         &
         {27.0}/{36.3}
         &
         {251.4}/{147.5}
         &
         {40.9}/{66.7}
         \\
         \rowcolor{colLoHa}
         LoHa: Hadamard
         &
         {\textbf{965.3}}/{\textbf{1079.7}}
         &
         {\textbf{38.5}}/{\textbf{43.0}}
         &
         {\textbf{332.5}}/{\textbf{219.6}}
         &
         {\textbf{80.4}}/{\textbf{82.9}}
         \\
         \rowcolor{colLoKr}
         LoKr: Kronecker
         &
         {844.4}/{449.3}
         &
         {37.4}/{40.7}
         &
         {312.5}/{62.5}
         &
         {74.8}/{70.2}
    \end{tabular}
    \vspace{-5pt}
    \label{tab:abl-operator}
\end{table}

Ablation of operators is conducted using Vicuna-7B and -13B under default settings with the best-tuned rank for each operator. The results reveal operator-specific strengths across benchmarks, indicating that the choice of low-rank operator is a promising avenue for future optimization.

\paragraph{Model scales and architectures.}
\begin{table}[t]
\centering
\caption{Performance across model scales and alignment methods. Full-parameter alignment serves as the baseline. MobileVLM-V2 results use rank 64 for all low-rank methods.}
\label{tab:scales}
\vspace{-12pt}
\tablestyle{15pt}{1.3}
\begin{tabular}{l l
    >{\columncolor{colFull}}r
    >{\columncolor{colLoRA}}r
    >{\columncolor{colLoKr}}r
    >{\columncolor{colLoHa}}r
}
{model} & {metric} &
\cellcolor{colFull}{Full} &
\cellcolor{colLoRA}{LoRA} &
\cellcolor{colLoKr}{LoKr} &
\cellcolor{colLoHa}{LoHa} \\
\hline
\multirow{2}{*}{1.4B}
  & MME-P  & 208.1 & 568.7 & \textbf{766.6} & 450.8 \\
  & POPE   &  2.5 & 51.5 & \textbf{72.8} & 63.1 \\
\hline
\multirow{2}{*}{2.7B}
  & MME-P  &  15.8 &  52.0 & \textbf{432.6} & 318.3 \\
  & POPE   &  0.5 & 15.0 & \textbf{71.3} & 66.1\\
\hline
\multirow{2}{*}{7B}
  & MME-P  & 521.0 & 551.8 & \textbf{825.1} & 608.3 \\
  & POPE   & 35.0 & 37.2 & \textbf{70.7} & 67.3 \\
\hline
\multirow{2}{*}{13B}
  & MME-P  & 236.6 & 288.0 &  78.5 & \textbf{621.4} \\
  & POPE   & \textbf{66.9} & 44.8 & 37.6 & 67.4 \\
\end{tabular}
\vspace{-10pt}
\end{table}

\begin{table}[t]
\centering
\caption{Performance of alignment methods on Qwen3. Low-rank alignment works on most of the benchmarks, with notable exceptions on POPE and MMMU-Pro where rank-constrained updates prove insufficient.}
\vspace{-10pt}
\label{tab:arc}
\tablestyle{2pt}{1.3}
\begin{tabular}{l
    >{\columncolor{colFull}}r
    >{\columncolor{colLoRA}}r
    >{\columncolor{colDelta}}r
    >{\columncolor{colFull}}r
    >{\columncolor{colLoRA}}r
    >{\columncolor{colDelta}}r
}
{metric} &
\cellcolor{colFull}full 8B &
\cellcolor{colLoRA}LoRA ($r\!=\!128$,  8B) &
\cellcolor{colDelta}$\Delta$ &
\cellcolor{colFull}full  14B&
\cellcolor{colLoRA}LoRA ($r\!=\!32$,  14B) &
\cellcolor{colDelta}$\Delta$ \\
\hline
GQA       &  28.4 & \textbf{37.2} & $+8.8$   &   6.6 & \textbf{20.2} & $+13.6$ \\
MMBench   &  18.1 & \textbf{48.7} & $+30.6$  &  12.0 & \textbf{45.5} & $+33.5$ \\
MME-P     & 536.0 & \textbf{752.3} & $+216.3$ & 526.3 & \textbf{988.9} & $+462.6$ \\
MME-R     & 212.1 & \textbf{242.1} & $+30.0$  & 234.3 & \textbf{308.6} & $+74.3$ \\
MMMU-Pro  & \textbf{12.1} &   5.7 & $-6.4$   & \textbf{11.8} & \textbf{11.8} & $0$ \\
POPE      & \textbf{92.7} &  25.1 & $-67.6$  &   n/a & \textbf{62.3} & n/a \\
ScienceQA &  63.7 & \textbf{71.3} & $+7.6$   &  59.3 & \textbf{71.9} & $+12.6$ \\
TextVQA   &   5.9 & \textbf{9.7}  & $+3.8$   &   4.2 & \textbf{10.2} & $+6.0$ \\
\end{tabular}
\vspace{-10pt}
\end{table}

As shown in Tables~\ref{tab:scales} and~\ref{tab:arc}, low-rank alignment consistently matches or outperforms full fine-tuning across scales from 1.4B to 14B, and this advantage generalizes to architecturally distinct models including Qwen3-8B/14B under a one-shot protocol. Two exceptions, POPE and MMMU-Pro, appear consistently across both model families, where higher-rank updates partially overwrite pre-trained knowledge structures. The rank dependence of these anomalies suggests that the low-rank constraint acts as a regularizer limiting deviation from the pre-trained weight manifold, a property that becomes increasingly important as rank grows.

\definecolor{colEncoder}{HTML}{E8F4FD}
\definecolor{colFreeze}{HTML}{EAF7EA}
\definecolor{colScope}{HTML}{FEF9E7}
\definecolor{colOverfit}{HTML}{FDEDEC}

\begin{table}[t]
\centering
\caption{Ablation studies on alignment design choices. All configurations use full fine-tuning in the instruction tuning stage with other default settings unchanged. Row colors denote ablation groups: \colorbox{colEncoder}{visual encoder}, \colorbox{colFreeze}{frozen vs.\ unfrozen encoder}, \colorbox{colScope}{low-rank adaptation scope}, and \colorbox{colOverfit}{overfitting control}.}
\vspace{-10pt}
\label{tab:ablation}
\tablestyle{5pt}{1.2}
\begin{tabular}{lrrrrrr}
bench. & GQA & MME-p & POPE & TextVQA & MMB-dev & SQA \\
\hline
\rowcolor{colEncoder}
dinov2-full      & 61.56 & 1503.69 & 86.86 & \textbf{51.56} & 72.51 & 74.27 \\
\rowcolor{colEncoder}
dinov2-LoRA    & 61.88 & 1481.45 & 86.52 & 55.19 & \textbf{74.66} & \textbf{76.15} \\
\rowcolor{colEncoder}
siglipv2-full    & 62.19 & 1406.61 & \textbf{88.24} & 29.58 & 72.68 & 73.42 \\
\rowcolor{colEncoder}
siglipv2-LoRA  & \textbf{62.73} & \textbf{1474.70} & 87.52 & 30.82 & 73.37 & 72.08 \\
\hline
\rowcolor{colFreeze}
frz.\ clip-LoRA       & 62.47 & \textbf{1511.61} & 87.52 & \textbf{44.73} & \textbf{72.41} & 71.42 \\
\rowcolor{colFreeze}
un-frz.\ clip-full      & 59.98 & 1444.91 & 85.79 & 40.18 & 71.74 & 72.53 \\
\rowcolor{colFreeze}
un-frz.\ clip-LoRA    & 60.18 & 1373.54 & 86.49 & 42.97 & 71.22 & \textbf{72.53} \\
\rowcolor{colFreeze}
un-frz.\ dinov2-full    & 61.78 & 1403.20 & 87.60 & 28.61 & 71.82 & 72.38 \\
\rowcolor{colFreeze}
un-frz.\ dinov2-LoRA  & \textbf{63.07} & 1481.98 & \textbf{88.18} & 37.27 & 76.03 & 72.04 \\
\hline
\rowcolor{colScope}
all (attn.\ + MLP)  & \textbf{62.47} & \textbf{1511.61} & 87.52 & \textbf{44.73} & 72.41 & 71.42 \\
\rowcolor{colScope}
MLP only            & 59.99 & 1384.71 & \textbf{87.95} & 41.29 & \textbf{72.77} & \textbf{75.41} \\
\rowcolor{colScope}
attn.\ only         & 62.26 & 1391.84 & 87.66 & 39.38 & 69.62 & 74.32 \\
\hline
\rowcolor{colOverfit}
full + weight decay  & 61.30 & 1432.68 & 86.21 & 43.55 & 70.34 & 70.61 \\
\rowcolor{colOverfit}
default (LoRA)     & \textbf{62.47} & \textbf{1511.61} & \textbf{87.52} & \textbf{44.73} & \textbf{72.41} & \textbf{71.42} \\
\hline
\end{tabular}
\vspace{-10pt}
\end{table}

\paragraph{Other ablation on default settings.} These ablations jointly validate the default configuration of our alignment pipeline across four dimensions.
(a) Visual encoder. Both DINOv2~\cite{oquab2023dinov2} and SigLIPv2~\cite{tschannen2025siglip} backbones benefit from low-rank alignment over full fine-tuning, with consistent gains on GQA and MMBench. The improvement is more pronounced under DINOv2, suggesting that the low-rank constraint interacts more favorably with dense spatial features than with the global token representations of SigLIPv2.
(b) Encoder freezing. Freezing the visual encoder during alignment consistently outperforms unfreezing across nearly all metrics, regardless of backbone or alignment method. Unfreezing introduces instability, particularly on TextVQA and MME-p, indicating that gradient updates to the encoder during alignment disrupt pre-trained visual representations rather than refining them.
(c) Adaptation scope. Applying low-rank adaptation jointly to both attention and MLP modules yields the best overall performance, with the full configuration outperforming MLP-only and attention-only variants on MME-p and TextVQA. This suggests that the two module types capture complementary aspects of visual-language alignment and should not be decoupled.
(d) Overfitting. Our default low-rank alignment matches or exceeds full fine-tuning with tuned weight decay across all six benchmarks, confirming that the performance advantage is not attributable to overfitting in the full fine-tuning baseline. The low-rank constraint is an effective overfitting regularizer.

\vspace{-3mm}
\section{Related Works}
\paragraph{Low-rank Adaptation and Vision-Language Models.}
Low-rank adaptation~\cite{hulora} introduces trainable low-rank matrices into frozen linear layers for parameter-efficient fine-tuning, with extensions including QLoRA~\cite{dettmers2024qlora}, LoHa~\cite{loha}, LoKr~\cite{lokr}, and further variants~\cite{liu2024dora, hayou2024lora+, meng2024pissa, xu2023qa, li2023loftq,zhou2025gsq}. On the VLM side, Flamingo~\cite{alayrac2022flamingo} pioneered visual-language integration, inspiring a broad family of models~\cite{li2023blip2, liu2023llava, zhu2023minigpt, chen2024internvl, chu2024mobilevlm2, yao2024minicpmv} that differ in scale, architecture, and alignment strategy. In this context, VLoRA~\cite{ma2024visual} suggests that cross-modal information transfer may be intrinsically low-rank, providing empirical motivation for applying low-rank adaptation in the alignment stage. However, whether such constraints introduce systematic implicit biases in visual token representations during alignment remains uncharacterized. Our work targets this gap: rather than proposing a new method, we analyze the alignment dynamics induced by LLM-side low-rank adaptation, a question orthogonal to connector design. Since many open-source VLMs release only trained weights without the full reproduction materials required to replicate their pretraining stage, we adopt a more general alignment pipeline that abstracts away model-specific details while remaining representative of mainstream practice, as evidenced by its continued use in recent popular models such as Kimi2.5-K2~\cite{kimik25}. Our simplified pipeline is therefore deliberately distinct from architecture-specific designs such as Qwen3-VL-style~\cite{bai2025qwen3} multi-layer visual injection (deepstack), which depends on proprietary training configurations beyond public reproduction, and we cannot reproduce the alignment stages of Qwen3-VL. Moreover, most current open-source models employ a ViT inherited from a previously aligned version rather than trained from unaligned ones, making it difficult to isolate for academic research.

\paragraph{Implicit Bias of Low-Rank Adaptation and Matrix Factorization.}
A classical line of work studies the implicit regularization of factorized parameterizations: gradient descent on matrix factorization is biased toward low-rank / minimal nuclear-norm solutions~\cite{gunasekar2017implicit}, an effect that deep factorization intensifies~\cite{arora2019implicit} and that need not be captured by any norm~\cite{razin2020implicit}; more recent analyses study LoRA from a parameter-efficiency or optimization-landscape standpoint~\cite{biderman2024lora,jang2024lora}. Our setting departs from this line in four respects: a non-linear LLM backbone, fine-tuning from a pretrained checkpoint, stochastic (noisy) dynamics, and a cross-entropy alignment objective. Consequently, it yields a \emph{flat-gradient, noise-robust subspace} bias (Sec.~\ref{sec:theory}) rather than the min-norm/low-rank bias of the classical setting; we expand on these four differences in Appendix~\ref{appdx:implicit-bias-comparison}.

\vspace{-2mm}
\section{Conclusion}
\vspace{-2mm}
This work investigates the implicit biases introduced by low-rank adaptation during vision-language alignment, a stage widely treated as requiring full-parameter updates. We show that low-rank alignment not only reduces training costs but consistently outperforms full fine-tuning across model scales from 1.4B to 14B parameters. Through behavioral, feature-level, and geometric analyzes, we identify three systematic biases: a shift from hallucinatory to conservative perception, preservation of per-token linear separability (in contrast to full alignment), and structurally homogeneous visual representations. These observations reveal that low-rank alignment retains general modality-specific knowledge during the alignment stage while deferring entity-specific fusion to subsequent instruction tuning. Two theorems ground this behavior in optimization theory, showing that low-rank gradient flow preferentially reinforces noise-robust feature directions and that steady-state parameters concentrate on flat, perturbation-robust subspaces. Extensive experiments across over 100 alignment configurations (spanning model scales, operator families and vision encoders) validate these findings and offer practical guidance for vision-language alignment.

\section*{Acknowledgements}
We thank the ECCV committee and reviewers for their careful reading and
constructive feedback. We are also grateful to colleagues for helpful discussions and to the
institutions that provided computational resources used in this study. The first author carried out and led this work as an independent researcher.

\bibliographystyle{splncs04}
\bibliography{main}
\newpage
\section*{Appendix}
\setcounter{section}{0}

\section{Implicit Biases: Low‑Rank Alignment}
\label{appdx:theorem}
\subsection{Preliminary: First-Order and Symbols}
Due to the high‑order tensor and Kronecker structures employed in our derivations, we adopt the Einstein summation convention by default. The analysis of the optimized feature $H$ is motivated by the Unconstrained Feature Model (UFM) under the universal approximation assumption and the practical large scale of both data and models.

\subsection*{Model Definition}
\[
    O^i = W_0{}^{i}{}_{j} \, H^{j} + b^{i} + l^{i}(H;\theta),
\]
where \(H^{j}\) is the input vector (dimension \(q\)), \(O^{i}\) is the output vector (dimension \(p\)), and \(\theta\) denotes, collectively, the parameters of \(l(H;\theta)\).

\subsection*{Symbols}
\begin{tabular}{ll}
\hline
Symbol & Meaning / Dimension \\ \hline
$O^i, H^j$ & Output/input vectors, $p$, $q$ \\
$O^{is}, H^{jt}$ & Kronecker Output/input, $p\times s$, $q\times t$ \\
$W_0^{\,i}{}_{j}, b^i$ & Base weight $p\times q$, bias $p$ \\
$A^{\,k}{}_{j}, B^{\,i}{}_{k}$ & Low-rank factors, $r\times q$, $p\times r$ \\
$B_1,B_2,A_1,A_2$ & Hadamard-case factors, $p\times r$, $r\times q$ \\
$C^{\,s}{}_{t}$ & Kronecker factor, $s\times t$ \\
$l^i(H;\theta)$ & Adaptation term with parameters $\theta$ \\
$F(O)$ & Scalar objective \\
$g^{(P)}_{ij}, g^{(Q)}_{ij}$ & Metric tensors; $g_{(P)}^{ij}, g_{(Q)}^{ij}$ inverses \\
$\delta_i^j$ & Kronecker delta \\
$*, \otimes$ & Hadamard, Kronecker products \\
$i,j,k$ & Output, input, latent indices \\
$m,n,s,t$ & Auxiliary summation indices \\
$M$ & Temporary variables for simplification \\
\hline
\end{tabular}

\subsubsection*{Index and Dimension Conventions}
\[
    i \in \{1,\dots,p\}, \quad j \in \{1,\dots,q\}, \quad k \in \{1,\dots,r\},
\]
and \(m,n,s,t\)\; are auxiliary summation indices.
All indices obey the Einstein summation convention: repeated upper and lower indices are summed.

\subsubsection*{Metric Tensors}
Let \(g^{(P)}_{ij}\) and \(g^{(Q)}_{ij}\) denote the metric tensors on output and input spaces respectively, with inverses \(g_{(P)}^{ij}\) and \(g_{(Q)}^{ij}\) satisfying
\[
    g^{(P)}_{\,i k}\;g_{(P)}^{\,k j} = \delta_i{}^{j}, \qquad
    g^{(Q)}_{\,i k}\;g_{(Q)}^{\,k j} = \delta_i{}^{j}.
\]
For covariant differentiation we write:
\[
    \frac{\partial F}{\partial H_{\,j}}
      = g^{(Q)}_{\,j m}\; \frac{\partial F}{\partial H^{\,m}}, \qquad
    \frac{\partial F}{\partial O_{\,i}}
      = g^{(P)}_{\,i m}\; \frac{\partial F}{\partial O^{\,m}}.
\]

\subsection*{General Chain Rule}
\[
    \frac{\partial F}{\partial H^{\,j}}
      = \frac{\partial F}{\partial O^{\,i}}\;\frac{\partial O^{\,i}}{\partial H^{\,j}}, \qquad
    \frac{\partial F}{\partial \theta}
      = \frac{\partial F}{\partial O^{\,i}}\;\frac{\partial O^{\,i}}{\partial \theta}.
\]
Since
\[
    O^{\,i} = W_0{}^{\,i}{}_{j} \, H^{\,j} + b^{\,i} + l^{\,i}(H;\theta),
\]
it follows that
\[
    \frac{\partial O^{\,i}}{\partial H^{\,j}}
      = W_0{}^{\,i}{}_{j} + \frac{\partial l^{\,i}}{\partial H^{\,j}}.
\]

\subsection*{Case 0 full: $l=0$ (Pure Linear Map)}

\paragraph{Model.}
\[
    O^i = W_0{}^i{}_j H^j + b^i
\]

\paragraph{Gradients.}
Let $G_i = \frac{\partial F}{\partial O^i}$. Then
\begin{align*}
    \frac{\partial F}{\partial H^k} &= G_i W_0{}^i{}_k, \\
    \frac{\partial F}{\partial W_0{}^i{}_j} &= G_i H^j, \quad
    \frac{\partial F}{\partial b^i} = G_i.
\end{align*}

\paragraph{First-order conditions.}
\[
\begin{aligned}
    G_i W_0{}^i{}_k &= 0, \\
    G_i H^j &= 0, \\
    G_i &= 0
\end{aligned}
\]

\subsection*{Case 1: LoRA $l = BAH$}

\paragraph{Model.}
\[
l^i = B^i{}_m A^m{}_j H^j = M^i{}_j H^j, \quad M^i{}_j = B^i{}_m A^m{}_j
\]
\[
O^i = W_0{}^i{}_j H^j + b^i + M^i{}_j H^j
\]

\paragraph{Gradients.}
\begin{align*}
    \frac{\partial F}{\partial H^k} &= G_i (W_0{}^i{}_k + B^i{}_m A^m{}_k), \\
    \frac{\partial F}{\partial B^i{}_m} &= G_i A^m{}_j H^j, \quad
    \frac{\partial F}{\partial A^m{}_j} = G_i B^i{}_m H^j
\end{align*}

\paragraph{First-order condition.}
\[
\begin{aligned}
    G_i(W_0{}^i{}_k + B^i{}_mA^m{}_k) &= 0, \\
    G_iA^m{}_jH^j &= 0, \\
    G_iB^i{}_mH^j &= 0
\end{aligned}
\]

\subsection*{Case 2 LoHa: $l = [(B_1A_1)*(B_2A_2)]H$ (Hadamard)}

\paragraph{Model.}
\begin{align*}
M_1^i{}_j &= B_1^i{}_m A_1^m{}_j, \quad
M_2^i{}_j = B_2^i{}_m A_2^m{}_j, \\
M^i{}_j &= M_1^i{}_j M_2^i{}_j, \quad
l^i = M^i{}_j H^j
\end{align*}

\paragraph{Gradients.}
\begin{align*}
\frac{\partial F}{\partial H^k} &= G_i(W_0{}^i{}_k + M^i{}_k), \\
\frac{\partial F}{\partial B_1^s{}_t} &= G_s A_1^t{}_j M_2^s{}_j H^j, \quad
\frac{\partial F}{\partial B_2^s{}_t} = G_s A_2^t{}_j M_1^s{}_j H^j, \\
\frac{\partial F}{\partial A_1^m{}_n} &= G_i B_1^i{}_m M_2^i{}_n H^n, \quad
\frac{\partial F}{\partial A_2^m{}_n} = G_i B_2^i{}_m M_1^i{}_n H^n
\end{align*}

\paragraph{First-order conditions.}
\[
\begin{aligned}
    G_i(W_0{}^i{}_k + M_1^i{}_k M_2^i{}_k) &= 0, \\
    G_i B_1^i{}_m A_1^m{}_j H^j &= 0, \\
    G_i B_2^i{}_m A_2^m{}_j H^j &= 0
\end{aligned}
\]

\subsection*{Case 3 LoKr: $l = [C \otimes (BA)]H$ (Kronecker)}

\paragraph{Model.}
\[
M^i{}_j = B^i{}_m A^m{}_j, \quad
l^{ip} = M^i{}_j C^p{}_q H^{jq}
\]

\paragraph{Gradients.}
\begin{align*}
\frac{\partial F}{\partial H^{kr}} &= G_{ip}(\delta^i_k\delta^p_r W_0^{ip} + M^i{}_k C^p{}_r), \\
\frac{\partial F}{\partial B^s{}_t} &= G_{sp} C^p{}_q A^t{}_j H^{jq}, \quad
\frac{\partial F}{\partial A^m{}_n} = G_{ip} C^p{}_q B^i{}_m H^{nq}, \\
\frac{\partial F}{\partial C^p{}_q} &= G_{ip} M^i{}_j H^{jq}
\end{align*}

\paragraph{First-order conditions.}
\[
\begin{aligned}
    G_{ip}(W_0^{ip}+M^i{}_jC^p{}_q) H^{jq} &= 0, \\
    G_{ip} A^m{}_j B^i{}_m H^{jq} &= 0, \\
    G_{ip} M^i{}_j H^{jq} &= 0
\end{aligned}
\]

\paragraph{Notes}
\begin{itemize}
    \item Einstein summation convention applies to all repeated indices.
    \item $\delta^i_j$ is the Kronecker delta.
    \item $g_{ij}$ and $h_{jk}$ are metric tensors for covariant/contravariant transformations.
    \item $M$, $M_1$, $M_2$ denote intermediate low-rank mappings.
\end{itemize}

\subsection{LoRA Rank-Weighted Smoothing under Noisy Features}
Consider a neural network with weights $W_{0,j}^{i}$ and a low-rank update parameterized by $A^p{}_j$ and $B^i{}_p$. Let the feature vector be $H^j$, corrupted by additive noise $\xi^j$:
\[
H^j = \bar H^j + \xi^j, \quad
\mathbb{E}[\xi^j] = 0, \quad
\mathbb{E}[\xi^j \xi^k] = \Sigma^{jk},
\]
where
\begin{itemize}
    \item $\bar H^j$ is the noise-free feature,
    \item $\xi^j$ is the additive noise,
    \item $\Sigma^{jk}$ is the noise covariance matrix.
\end{itemize}

Let the total loss function be $F = F(W,H)$. The LoRA gradient flow for $A$ is
\[
\dot A^p{}_j = -\eta B^i{}_p \frac{\partial F}{\partial W_{0}^{i}{}_j},
\]
where $\eta>0$ is the learning rate.

\paragraph{Taylor expansion of the gradient}
Expanding the gradient of $F$ w.r.t $W$ around the noise-free feature $\bar H$ using a multivariate Taylor expansion:
\begin{align*}
\frac{\partial F}{\partial W_{0}^{i}{}_j}(H)
=& \frac{\partial F}{\partial W_{0}^{i}{}_j}(\bar H)
+ \frac{\partial^2 F}{\partial W_{0}^{i}{}_j \partial H^k} \xi^k
\\
&+ \frac{1}{2} \frac{\partial^3 F}{\partial W_{0}^{i}{}_j \partial H^k \partial H^l} \xi^k \xi^l
+ \mathcal{O}(\|\xi\|^3),
\end{align*}
where indices follow Einstein summation convention.

\paragraph{Expectation over noise}
Taking expectation over the noise $\xi$:
\begin{align*}
\mathbb{E}_\xi\left[ \frac{\partial F}{\partial W_{0}^{i}{}_j} \right]
&\simeq \frac{\partial F}{\partial W_{0}^{i}{}_j}(\bar H)
+ \frac{1}{2} \frac{\partial^3 F}{\partial W_{0}^{i}{}_j \partial H^k \partial H^l} \mathbb{E}[\xi^k \xi^l]
\\
&
= \frac{\partial F}{\partial W_{0}^{i}{}_j}(\bar H)
+ \frac{1}{2} \frac{\partial^3 F}{\partial W_{0}^{i}{}_j \partial H^k \partial H^l} \Sigma^{kl}.
\end{align*}

\paragraph{Substitute into LoRA gradient flow}
Substituting the expected gradient into the LoRA gradient flow:
\[
\dot A^p{}_j \simeq -\eta B^i{}_p
\left(
\frac{\partial F}{\partial W_{0}^{i}{}_j}(\bar H)
+ \frac{1}{2} \frac{\partial^3 F}{\partial W_{0}^{i}{}_j \partial H^k \partial H^l} \Sigma^{kl}
\right).
\]

\subsection{Implicit Biases of Low-Rank Alignment: Gradient Flow}
\begin{table*}[ht]
\tablestyle{1pt}{1.3}
\centering
\caption{Implicit biases summary and comparison: LoRA, LoKr, LoHa}
\begin{tabular}{cccc}
\hline
Property & LoRA & LoKr & LoHa \\
\hline
Grad. Flow Subspace & Rank-$r$ ($B,A$) & Rank-$r$ ($C\otimes BA$) & Rank-$r_1 \times r_2$ ($(B_1A_1)*(B_2A_2)$) \\
Flatness Bias & Along $BA$ & Along $C\otimes BA$ & Along $(B_1A_1)*(B_2A_2)$ \\
Noise Smoothing & Along $H$ & Through $C\otimes BA$ & Along Hadamard subspace \\
Scale Invariance & Linear in $H$ & Linear in $H$ & Linear in $H$ \\
\hline
\end{tabular}
\label{tab:implicit_bias_column}
\end{table*}

\paragraph{Case 1 LoRA: $l=BAH$}

Consider the LoRA forward mapping:
\begin{equation*}
O^i = (W_0^{i}{}_{j} + B^{i}{}_{p} A^{p}{}_{j}) H^j,
\end{equation*}
with a total loss function
\begin{equation*}
F = \mathbb{E}_{(x,y)\sim \mathcal{D}} \left[ \mathcal{L}(O(x),y) \right].
\end{equation*}

Assume the feature has additive noise:
\begin{equation*}
H^j = \bar H^j + \xi^j, \quad
\mathbb{E}[\xi^j]=0, \quad
\mathbb{E}[\xi^j\xi^k]=\Sigma^{jk}.
\end{equation*}

\paragraph{Gradient flow dynamics.}
The gradient flow dynamics for LoRA parameters are:
\begin{align*}
\dot{A}^{p}{}_{j} &= -\eta \frac{\partial F}{\partial A^{p}{}_{j}}
= -\eta B^{i}{}_{p} \frac{\partial F}{\partial W_{0}^{i}{}_{j}}, \\
\dot{B}^{i}{}_{p} &= -\eta \frac{\partial F}{\partial B^{i}{}_{p}}
= -\eta \frac{\partial F}{\partial W_{0}^{i}{}_{j}} A^{p}{}_{j},
\end{align*}
with
\begin{equation*}
\frac{\partial F}{\partial W_{0}^{i}{}_{j}} = \frac{\partial F}{\partial O^i} H^j.
\end{equation*}

Define the gradient projection:
\begin{equation*}
G^{ij} := \mathbb{E}\left[ \frac{\partial F}{\partial O^i} H^j \right].
\end{equation*}

Thus, the expected gradient flow becomes:
\begin{align*}
\dot{A}^{p}{}_{j} &= -\eta B^{i}{}_{p} G^{ij}, \\
\dot{B}^{i}{}_{p} &= -\eta G^{ij} A^{p}{}_{j}.
\end{align*}

This shows that the gradient flow is restricted to the rank-$r$ subspace spanned by $B$ and $A$, \textit{w.r.t} the range of $p$.

\paragraph{Noise-weighted smoothing.}
Expanding the gradient under feature noise $\xi$:
\begin{equation*}
\mathbb{E}_\xi\left[ \frac{\partial F}{\partial W_{0}^{i}{}_{j}} \right]
\simeq \frac{\partial F}{\partial W_{0}^{i}{}_{j}}(\bar H)
+ \frac{1}{2} \frac{\partial^3 F}{\partial W_{0}^{i}{}_{j} \partial H^k \partial H^l} \Sigma^{kl}.
\end{equation*}

Substituting into the gradient flow:
\begin{equation*}
\dot{A}^{p}{}_{j} \approx -\eta B^{i}{}_{p}
\left(
\frac{\partial F}{\partial W_{0}^{i}{}_{j}} + \frac{1}{2} \frac{\partial^3 F}{\partial W_{0}^{i}{}_{j} \partial H^k \partial H^l} \Sigma^{kl}
\right),
\end{equation*}

showing an implicit smoothing effect along noisy feature directions.

\paragraph{Scale invariance.}
Under feature scaling $H^j \to \alpha H^j$, the gradient projection scales linearly:
\begin{equation*}
G^{ij} \to \alpha G^{ij}, \quad
\dot{A}^{p}{}_{j} \to \alpha \dot{A}^{p}{}_{j}, \quad
\dot{B}^{i}{}_{p} \to \alpha \dot{B}^{i}{}_{p}.
\end{equation*}

This indicates that the implicit flatness preference is invariant to feature scale.

\paragraph{Fokker--Planck approximation and steady-state distribution.}
Small-step SGD or noisy gradient flow can be approximated as an SDE:
\begin{equation*}
d\theta^\alpha = -\frac{\partial F}{\partial \theta^\alpha} dt + \sqrt{\eta}\, C^\alpha{}_\mu dw^\mu_t,
\end{equation*}
with noise covariance
\begin{equation*}
D_{\alpha\beta} = C^\gamma{}_\alpha C^\gamma{}_\beta.
\end{equation*}

The corresponding Fokker--Planck equation is:
\begin{equation*}
\partial_t p(\theta,t) = \partial_\alpha \left[ \partial^\alpha F \, p + \frac{1}{2} \partial_\beta(D^{\beta\alpha} p) \right] + \frac{1}{2} \partial_\alpha \partial_\beta (D^{\alpha\beta} p).
\end{equation*}

Assuming no-flux boundary conditions, the steady-state distribution is approximately:
\begin{equation*}
p_{\rm stat}(\theta) \propto \exp\left(-\frac{2}{\eta} F(\theta)\right) \det(D(\theta))^{-1/2}.
\end{equation*}

\paragraph{LoRA: low-rank subspace marginalization.}

Define the effective LoRA subspace:
\begin{equation*}
M = BA.
\end{equation*}

Then the steady-state marginal over $M$ is:
\begin{align*}
p_{\rm stat}(M) \propto \exp&\left(-\frac{2}{\eta} F(M)\right)
\\
&\underbrace{\int_{BA=M} \det D(B,A)^{-1/2} d\mu(B,A)}_{=: V(M)}.
\end{align*}

\paragraph{Effective Loss Incorporating Flatness and Noise}

Define the effective loss in the low-rank subspace:
\begin{align*}
F_{\rm eff}(M) & = F(M) + \frac{\eta}{4} \log \det \widetilde D(M) \\
&+ \frac{1}{2} \Sigma^{jk}
\frac{\partial^2 F}{\partial O^i \partial O^m} (W_0^i{}_j + M^i{}_j)(W_0^m{}_k + M^m{}_k),
\end{align*}
where the first term is the original loss, the second term encodes gradient-noise-induced flatness, and the third term encodes feature-noise-weighted smoothing.

\paragraph{Case 2 LoHa: $l=[(B_1A_1)*(B_2A_2)]H$}

Consider the LoHa forward mapping:
\begin{equation*}
O^i = (B_1^i{}_p A_1^p{}_j) * (B_2^i{}_q A_2^q{}_j) H^j,
\end{equation*}
with a total loss function
\begin{equation*}
F = \mathbb{E}_{(x,y)\sim \mathcal{D}} \left[ \mathcal{L}(O(x),y) \right].
\end{equation*}

Assume additive feature noise:
\begin{equation*}
H^j = \bar H^j + \xi^j, \quad
\mathbb{E}[\xi^j]=0, \quad
\mathbb{E}[\xi^j\xi^k]=\Sigma^{jk}.
\end{equation*}

\paragraph{Gradient flow dynamics.}
\begin{align*}
\dot{A}_1^p{}_j &= (B_2^i{}_q A_2^q{}_j) \frac{\partial F}{\partial O^i} H^j B_1^i{}_p,
\\
\dot{B}_1^i{}_p &= ((B_2^i{}_q A_2^q{}_j) \frac{\partial F}{\partial O^i} H^j) A_1^p{}_j, \\
\dot{A}_2^q{}_j &= (B_1^i{}_p A_1^p{}_j) \frac{\partial F}{\partial O^i} H^j B_2^i{}_q,
\\
\dot{B}_2^i{}_q &= ((B_1^i{}_p A_1^p{}_j) \frac{\partial F}{\partial O^i} H^j) A_2^q{}_j.
\end{align*}

Define gradient projection:
\begin{equation*}
G^{ij} := \mathbb{E}\left[ \frac{\partial F}{\partial O^i} H^j \right].
\end{equation*}

\paragraph{Noise-weighted smoothing.}
\begin{align*}
\mathbb{E}_\xi&\left[ \frac{\partial F}{\partial O^i} (B_1A_1 * B_2A_2) H^j \right]
\simeq \frac{\partial F}{\partial O^i} (B_1A_1 * B_2A_2) \bar H^j
\\
&+ \frac{1}{2} \frac{\partial^3 F}{\partial O^i \partial H^m \partial H^n} (B_1A_1 * B_2A_2) \Sigma^{mn}.
\end{align*}

\paragraph{Scale invariance.}
Under $H^j \to \alpha H^j$:
\begin{equation*}
G^{ij} \to \alpha G^{ij}, \quad
\dot{A}_{1,2}, \dot{B}_{1,2} \to \alpha \dot{A}_{1,2}, \alpha \dot{B}_{1,2}.
\end{equation*}

\paragraph{Fokker--Planck and steady-state.}
\begin{equation*}
d\theta^\alpha = -\frac{\partial F}{\partial \theta^\alpha} dt + \sqrt{\eta}\, C^\alpha{}_\mu dw^\mu_t, \quad
D_{\alpha\beta} = C^\gamma{}_\alpha C^\gamma{}_\beta,
\end{equation*}
\begin{equation*}
p_{\rm stat}(\theta) \propto \exp\left(-\frac{2}{\eta} F(\theta)\right) \det(D(\theta))^{-1/2}.
\end{equation*}

\paragraph{Effective Loss in LoHa subspace.}
\begin{equation*}
M = (B_1A_1) * (B_2A_2),
\end{equation*}
\begin{align*}
F_{\rm eff}(M) & = F(M) + \frac{\eta}{4} \log \det \widetilde D(M) \\
&+ \frac{1}{2} \Sigma^{jk}
\frac{\partial^2 F}{\partial O^i \partial O^m} (W_0^i{}_j + M^i{}_j)(W_0^m{}_k + M^m{}_k).
\end{align*}

\paragraph{Case 3 LoKr: $l=[C\otimes (BA)]H$}

Consider the LoKr forward mapping:
\begin{equation*}
O^i = C^i{}_k (B^k{}_p A^p{}_j) H^j,
\end{equation*}
with a total loss function
\begin{equation*}
F = \mathbb{E}_{(x,y)\sim \mathcal{D}} \left[ \mathcal{L}(O(x),y) \right].
\end{equation*}

Assume the feature has additive noise:
\begin{equation*}
H^j = \bar H^j + \xi^j, \quad
\mathbb{E}[\xi^j]=0, \quad
\mathbb{E}[\xi^j\xi^k]=\Sigma^{jk}.
\end{equation*}

\paragraph{Gradient flow dynamics.}
The gradient flow for LoKr parameters:
\begin{align*}
\dot{A}^{p}{}_{j} &= -\eta B^k{}_p C^i{}_k \frac{\partial F}{\partial O^i} H^j, \\
\dot{B}^{k}{}_{p} &= -\eta C^i{}_k \frac{\partial F}{\partial O^i} H^j A^p{}_j, \\
\dot{C}^{i}{}_{k} &= -\eta \frac{\partial F}{\partial O^i} (B^k{}_p A^p{}_j H^j).
\end{align*}

Define the gradient projection:
\begin{equation*}
G^{ij} := \mathbb{E}\left[ \frac{\partial F}{\partial O^i} H^j \right].
\end{equation*}

Expected gradient flow:
\begin{align*}
\dot{A}^{p}{}_{j} &= -\eta B^k{}_p C^i{}_k G^{ij}, \\
\dot{B}^{k}{}_{p} &= -\eta C^i{}_k G^{ij} A^p{}_j, \\
\dot{C}^{i}{}_{k} &= -\eta G^{ij} B^k{}_p A^p{}_j.
\end{align*}

\paragraph{Noise-weighted smoothing.}
\begin{align*}
\mathbb{E}_\xi\left[ \frac{\partial F}{\partial O^i} (B^k{}_p A^p{}_j) H^j \right]
&\simeq \frac{\partial F}{\partial O^i} (B^k{}_p A^p{}_j) \bar H^j
\\
&+ \frac{1}{2} \frac{\partial^3 F}{\partial O^i \partial H^m \partial H^n} (B^k{}_p A^p{}_j) \Sigma^{mn}.
\end{align*}

\paragraph{Scale invariance.}
Under $H^j \to \alpha H^j$:
\begin{align*}
G^{ij} \to \alpha G^{ij}, \quad
\dot{A}^{p}{}_{j} \to \alpha \dot{A}^{p}{}_{j},
\\
\dot{B}^{k}{}_{p} \to \alpha \dot{B}^{k}{}_{p}, \quad
\dot{C}^{i}{}_{k} \to \alpha \dot{C}^{i}{}_{k}.
\end{align*}

\paragraph{Fokker--Planck approximation and steady-state distribution.}
\begin{equation*}
d\theta^\alpha = -\frac{\partial F}{\partial \theta^\alpha} dt + \sqrt{\eta}\, C^\alpha{}_\mu dw^\mu_t,
\quad
D_{\alpha\beta} = C^\gamma{}_\alpha C^\gamma{}_\beta,
\end{equation*}
\begin{equation*}
\partial_t p(\theta,t) = \partial_\alpha \left[ \partial^\alpha F \, p + \frac{1}{2} \partial_\beta(D^{\beta\alpha} p) \right] + \frac{1}{2} \partial_\alpha \partial_\beta (D^{\alpha\beta} p),
\end{equation*}
\begin{equation*}
p_{\rm stat}(\theta) \propto \exp\left(-\frac{2}{\eta} F(\theta)\right) \det(D(\theta))^{-1/2}.
\end{equation*}

\paragraph{Effective Loss in LoKr subspace.}
\begin{equation*}
M = C \otimes (B A),
\end{equation*}
\begin{align*}
F_{\rm eff}(M) & = F(M) + \frac{\eta}{4} \log \det \widetilde D(M) \\
&+ \frac{1}{2} \Sigma^{jk}
\frac{\partial^2 F}{\partial O^i \partial O^m} (W_0^i{}_j + M^i{}_j)(W_0^m{}_k + M^m{}_k).
\end{align*}

\subsection{Relation to Classical Implicit-Bias Results}
\label{appdx:implicit-bias-comparison}
A classical line of work studies the implicit regularization of factorized parameterizations: gradient descent on matrix factorization is biased toward low-rank / minimal nuclear-norm solutions~\cite{gunasekar2017implicit}, an effect that deep factorization intensifies~\cite{arora2019implicit} and that need not be captured by any norm~\cite{razin2020implicit}. More recent analyses study LoRA from a parameter-efficiency or optimization-landscape standpoint~\cite{biderman2024lora,jang2024lora}. Our analysis differs along four axes that together yield a qualitatively different conclusion: (i)~\emph{non-linear backbone}: prior results target linear or deep-linear networks, whereas we analyze a factorized update acting on a large non-linear LLM; (ii)~\emph{fine-tuning rather than training from scratch}: we study adaptation from a pretrained checkpoint $W_0$, so the relevant bias concerns deviation from a pretrained weight manifold rather than convergence from small initialization; (iii)~\emph{stochastic dynamics}: our account explicitly incorporates gradient noise instead of assuming deterministic, full-batch flow; and (iv)~\emph{alignment objective}: we consider a cross-entropy alignment loss rather than squared loss. Under these conditions, our theorems characterize a \emph{flat-gradient, noise-robust subspace} bias (Sec.~\ref{sec:theory}), distinct from the min-norm/low-rank bias of the classical setting, and it is this bias that underlies the structure-preserving behavior observed empirically.

\section{More experimental results.}
\label{appdx:extra_exp}
\subsection{Statement and Limitations}
We state the boundaries of our claims explicitly. \textbf{(i) Controlled alignment pipeline.} To isolate alignment-stage effects and keep attribution clean, we deliberately study a minimal pipeline rather than any single proprietary, architecture-specific recipe; our claims are correspondingly scoped to the alignment stage. Within this scope the observed bias is not a single-setup artifact: it holds consistently across model scales (1.4B--14B), three low-rank operator families, multiple vision encoders, and the architecturally distinct Qwen3 family. Compute constraints further restrict our finest-grained probing to smaller models and to already-post-trained large models. \textbf{(ii) Explanatory theory.} Our theorems are intended as a mechanistic explanation rather than a quantitative bound: they adopt idealized assumptions (an unconstrained-feature view and a small-step, noisy-gradient regime), in the spirit of standard NTK- and neural-collapse-style analyses. Their role is to predict the \emph{direction} of the observed biases (conservative behavior, homogeneous manifolds, and preserved linear separability), which the experiments corroborate, not to certify exact training trajectories. \textbf{(iii) Associational evidence.} We phrase the link between the feature-level and geometric signatures and downstream gains associationally rather than causally. This is deliberate: the support is a rank-controlled chain in which a single intervention (the low-rank constraint) produces parallel, monotone shifts in per-token linear separability, angular and manifold geometry, Type-I/II behavior, and downstream accuracy, consistently across three operator families and five model scales; we present this as strong, controlled association and do not overstate it as proof of causation. \textbf{(iv) Boundary cases.} The few benchmarks on which low-rank does not lead (\eg, POPE and MMMU-Pro at certain ranks) are consistent with, and predicted by, our regularization account (Sec.~\ref{sec:ablation}): they delimit the regime in which the implicit bias is beneficial rather than contradict it.

\subsection{Why Not Constrain the Connector?}
We apply low-rank adaptation to the LLM rather than to the vision--language connector (MLP adapter) by design. The connector holds only a tiny fraction of trainable parameters (\textless0.3\%), so a low-rank constraint there brings negligible compute or memory savings. More importantly, the connector is the sole bottleneck through which \emph{all} visual tokens pass; forcing it to be low-rank would project every visual feature into one shared low-rank subspace and discard token-specific structure, the opposite of the selective, structure-preserving regularization that LLM-side low-rank provides. We therefore keep the connector at full rank by default and study low-rank adaptation on the LLM.

\begin{table}[ht]
    \centering
    \tablestyle{4pt}{1.3}
    \caption{Effect of extended instruction datasets on full vs.\ low-rank aligned models (MobileVLM-2.7B). $\Delta$zero and $\Delta$Ex.\ denote gains under zero-shot and extended finetuning settings. Boost ratio $= \Delta\mathrm{Ex.}/\Delta\mathrm{zero} - 1$. Low-rank results use the best configuration among LoRA/LoHa/LoKr at rank 32/64/128.}
    \label{tab:extended_dataset}
    \begin{tabular}{lrrrrrr}
    metric & full & low-rank & $\Delta$zero & $\Delta$Ex.:full & $\Delta$Ex.:low-rank & boost (\%) \\
    \hline
    GQA       & 11.6 & 42.4 & 30.8$\uparrow$ & 13.7 & 3.2  & -34.3 \\
    MMBench   & 0.1  & 3.6  & 3.4$\uparrow$  & 2.7  & 30.3 & 805.2 \\
    MME-P     & 15.7 & 668.0 & 652.2$\uparrow$ & 148.7 & 406.4 & 39.6 \\
    MME-R     & 40.4 & 201.4 & 161.1$\uparrow$ & -19.6 & -7.5 & 7.5 \\
    MMMU-Pro  & 0.0  & 0.2  & 0.2$\uparrow$  & 3.8  & 5.7  & 824.7 \\
    POPE      & 0.5  & 79.4 & 78.8$\uparrow$ & 1.5  & -9.7  & -14.2 \\
    ScienceQA & 1.1  & 23.4 & 22.3$\uparrow$ & 5.9  & 29.2 & 104.1 \\
    TextVQA   & 0.6  & 20.5 & 19.8$\uparrow$ & 6.2  & 1.4  & -23.9 \\
    \end{tabular}
\end{table}

\subsection{Extended finetuning dataset.}
In this work, we primarily focus on zero-shot capabilities. We further investigate how efficiently a model with minimal domain knowledge or limited reasoning ability can absorb new information under both low-rank and full alignment settings, as shown in Table~\ref{tab:extended_dataset}. The largest model that is appropriate for this setting is MobileVLM-2.7B. GQA and TextVQA represent previously seen knowledge, whereas MMBench and ScienceQA correspond to novel sub-domain knowledge. We additionally track changes in perception and reasoning on MME and assess whether the model retains its zero-shot image reasoning ability on MMMU-Pro. POPE is a benchmark for visual perception hallucination.

\paragraph{Analyses.}
We observe limited gains from previously seen knowledge, such as GQA and TextVQA, although improvements are still present. In contrast, low-rank alignment yields pronounced gains on unseen, knowledge-intensive, or comprehensive benchmarks such as ScienceQA and MMBench. This further supports our hypothesis: low-rank alignment preserves more general knowledge and leaves sufficient capacity for subsequent entity-level construction.
We also find that new capabilities emerge on unseen tasks such as MMMU-Pro Vision, where accuracy rises from near-zero to meaningful levels, with low-rank alignment exhibiting stronger emergent behavior.
On the negative side, both reasoning performance and visual hallucination robustness degrade. The trade-off, however, is an improvement in perceptual ability driven by the increased infusion of visual knowledge.

\begin{table}[ht]
    \centering
    \tablestyle{4pt}{1.3}
    \caption{Comparison of full vs.\ low-rank alignment under full-parameter (F.F.T.) and default (LoRA, rank 128) instruction tuning. Model is MobileVLM-1.4B.}
    \label{tab:full-FT}
    \begin{tabular}{lrrrrrr}
    metric & full & low-rank & $\Delta$default & full F.F.T. & low-rank F.F.T. & $\Delta$FFT \\
    \hline
    GQA       & 18.9 & 27.7  & 8.8$\uparrow$   & 47.8 & 51.9 & 4.1$\uparrow$ \\
    MMBench   & 0.0  & 0.6   & 0.6$\uparrow$   & 2.1  & 9.2  & 7.1$\uparrow$ \\
    MME-P     & 208.1 & 766.6 & 558.5$\uparrow$ & 736.9 & 743.6 & 6.7$\uparrow$ \\
    MME-R     & 22.9 & 196.4 & 173.6$\uparrow$ & 80.4  & 140.0 & 59.6$\uparrow$ \\
    MMMU-Pro  & 1.0  & 4.4   & 3.5$\uparrow$   & 0.2  & 0.5  & 0.3$\uparrow$ \\
    POPE      & 2.5  & 72.8  & 70.3$\uparrow$  & 27.6 & 58.2 & 30.6$\uparrow$ \\
    ScienceQA & 0.1  & 16.4  & 16.2$\uparrow$  & 11.0 & 25.1 & 14.1$\uparrow$ \\
    TextVQA   & 4.2  & 8.1   & 3.8$\uparrow$   & 24.5 & 18.1 & -6.4 \\
    \end{tabular}
\end{table}

\subsection{Full parameter instruction tuning.}
This section examines the performance headroom of low-rank aligned models under different fine-tuning regimes, including full-parameter tuning. We observe that the smaller model, MobileVLM-1.4B, exhibits far fewer metric collapses when trained with low-rank alignment in Table~\ref{tab:full-FT}. Even among models that do not collapse, low-rank alignment consistently yields greater robustness.

\subsection{Linear separability of vision tower's feature space.}
\begin{figure*}[t]
    \centering
    \includegraphics[width=0.32\linewidth]{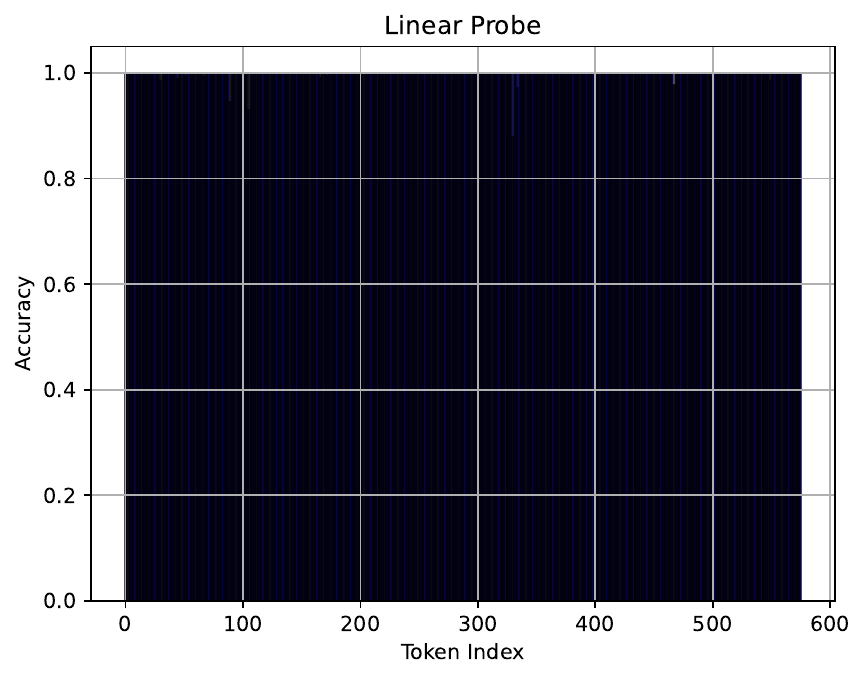}
    \includegraphics[width=0.32\linewidth]{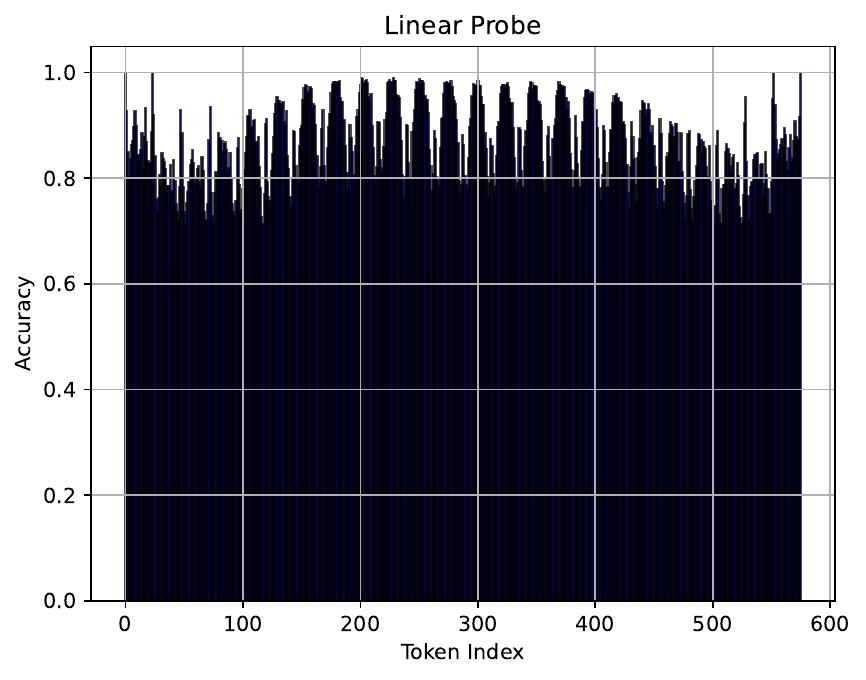}
    \includegraphics[width=0.32\linewidth]{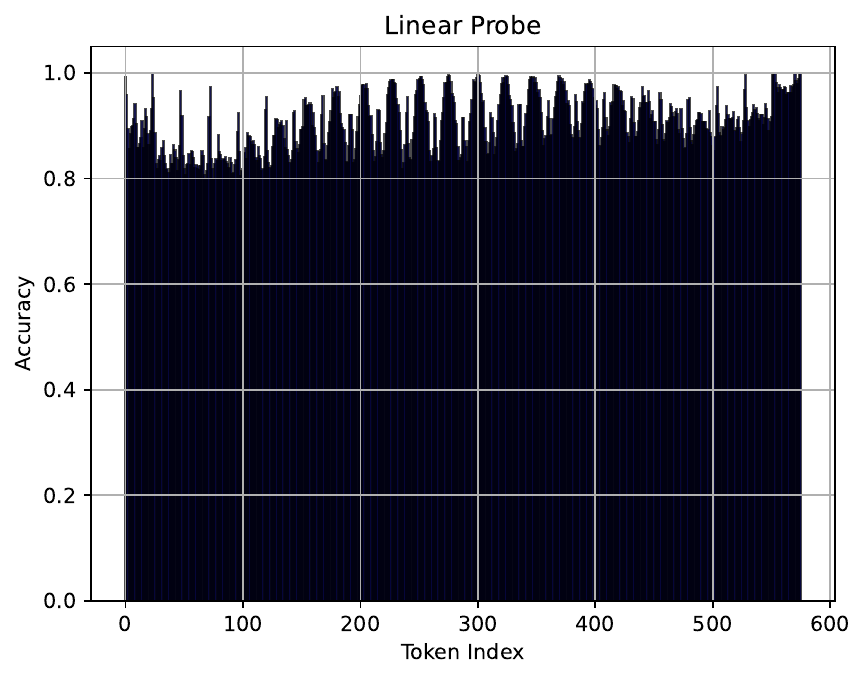}
    \begin{center}
        ImageNet (CLIP)
        \hspace{0.14\linewidth}
        STF-dogs (CLIP)
        \hspace{0.14\linewidth}
        CUB (CLIP)
    \end{center}
    \includegraphics[width=0.32\linewidth]{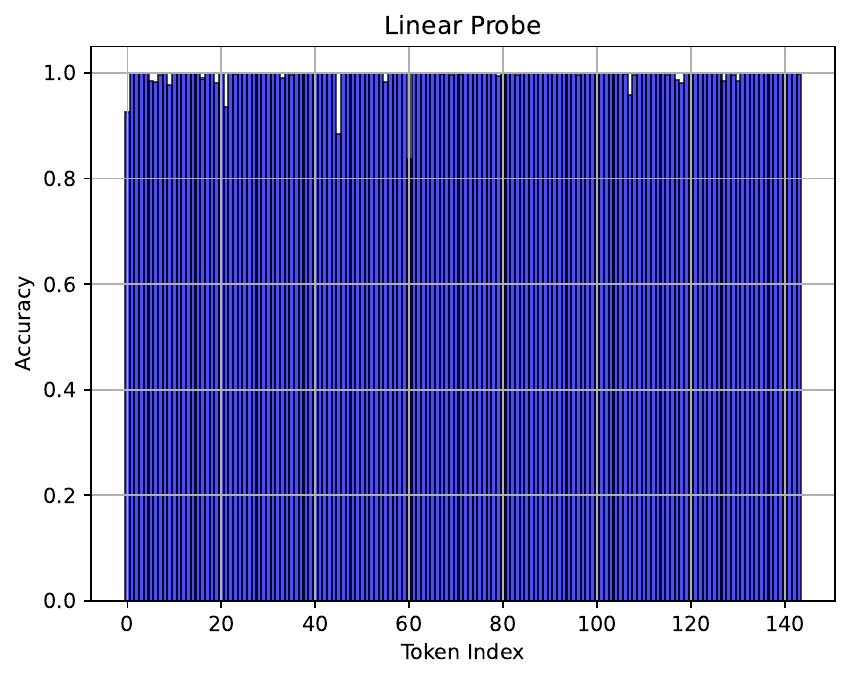}
    \includegraphics[width=0.32\linewidth]{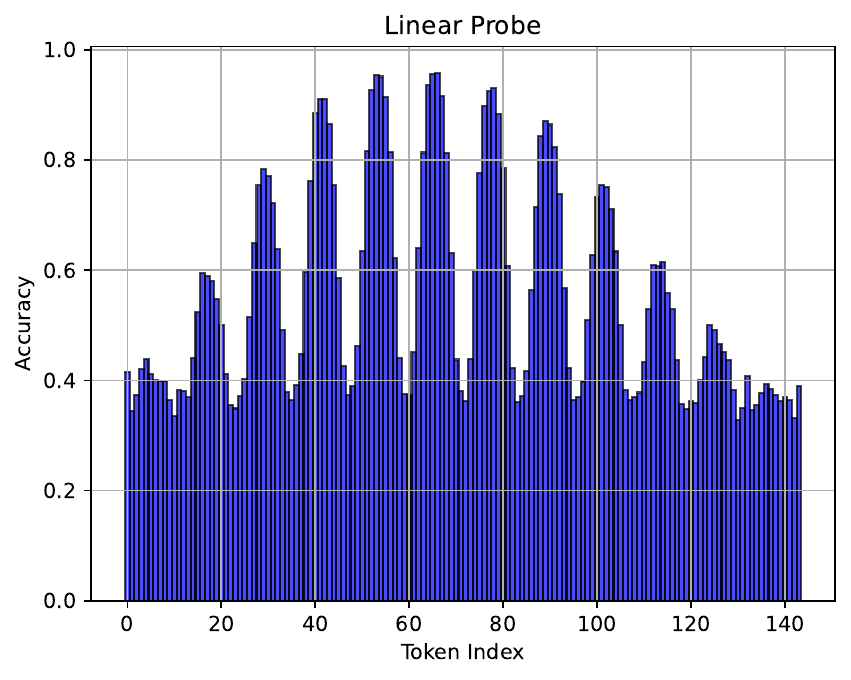}
    \includegraphics[width=0.32\linewidth]{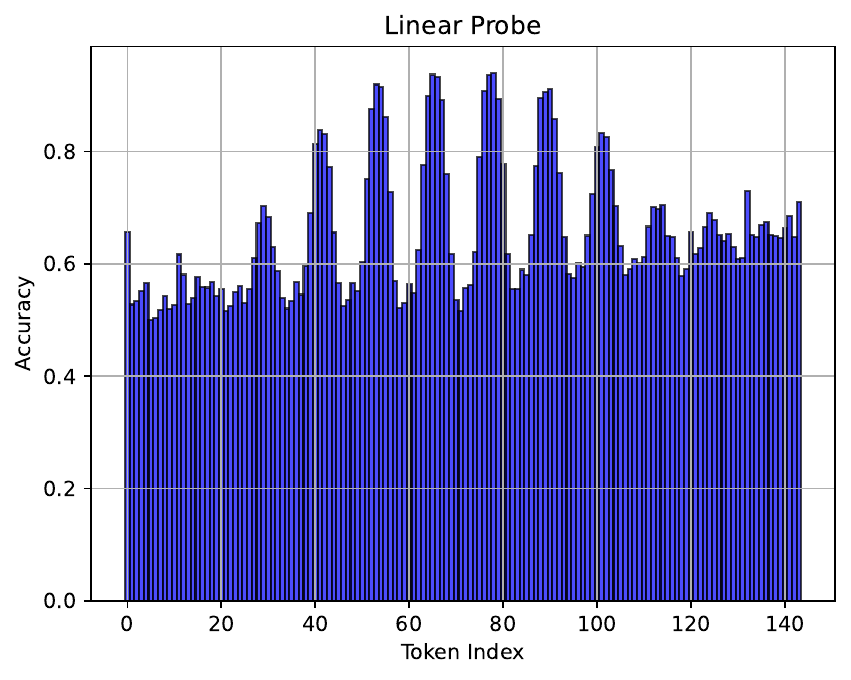}

        \begin{center}
        ImageNet (full)
        \hspace{0.16\linewidth}
        STF-dogs (full)
        \hspace{0.16\linewidth}
        CUB (full)
    \end{center}
    \includegraphics[width=0.32\linewidth]{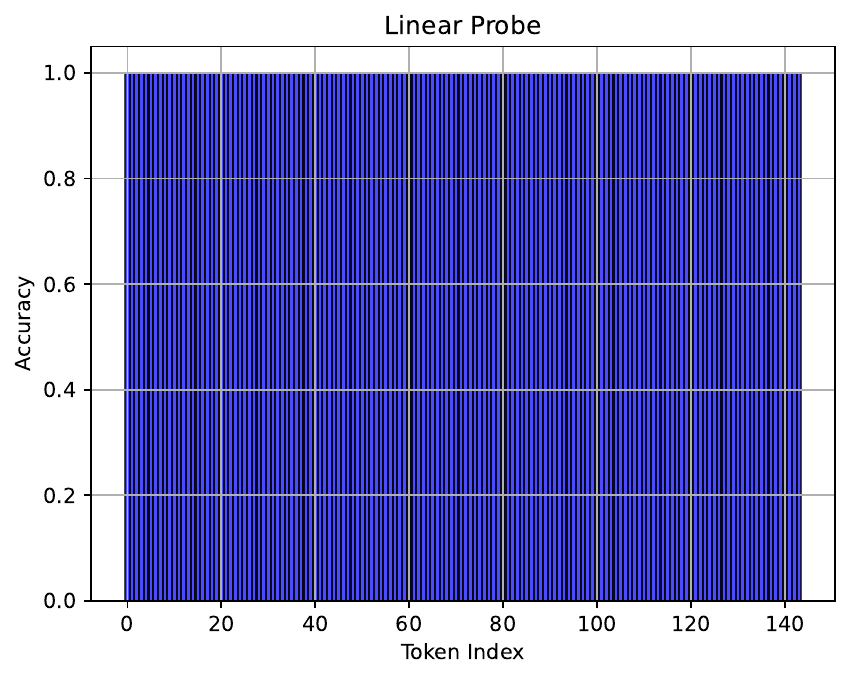}
    \includegraphics[width=0.32\linewidth]{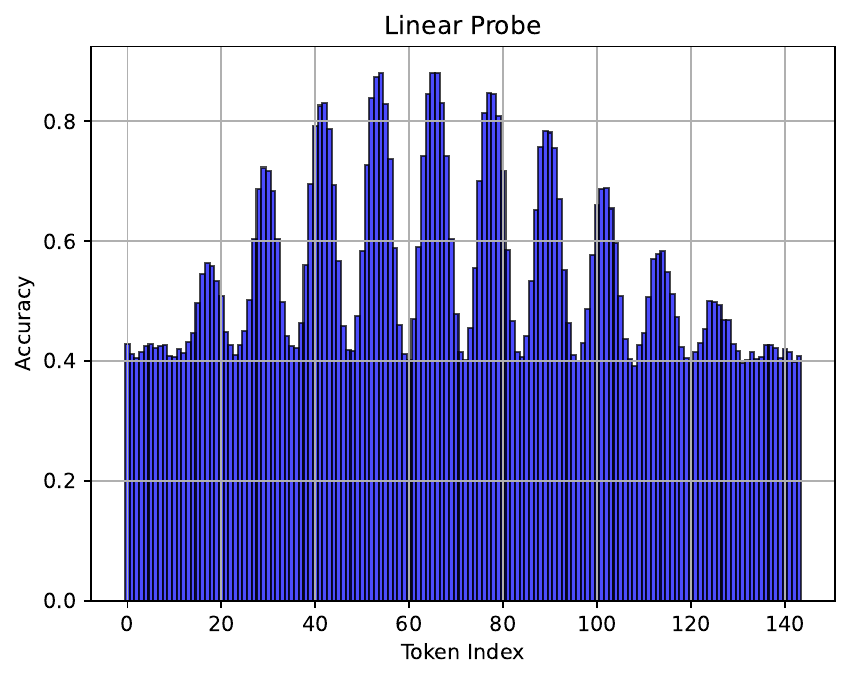}
    \includegraphics[width=0.32\linewidth]{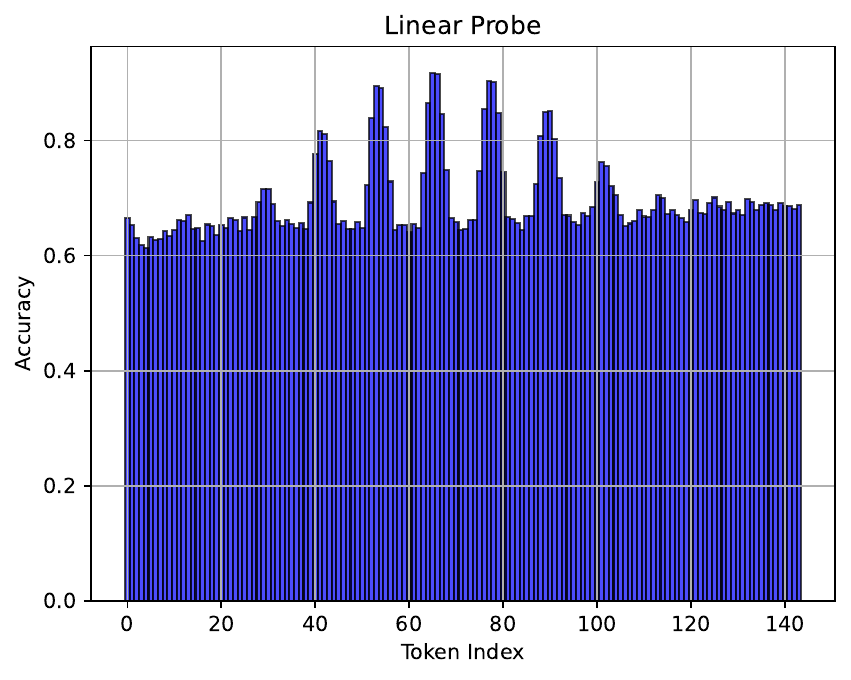}

        \begin{center}
        ImageNet (LoRA)
        \hspace{0.14\linewidth}
        STF-dogs (LoRA)
        \hspace{0.14\linewidth}
        CUB (LoRA)
    \end{center}

    \caption{CLIP, the vision tower of related VLMs, originally has great general and fine-grained knowledge structures in each token.
    Notation CLIP means features in original CLIP of 336$\times$336 resolutions and 14 patch.
    Notations full/LoRA mean features in each vision token from full-parameter/LoRA pretrained VLMs. Fine-grained classification datasets: STF-dogs and CUB.}
    \label{fig:CLIP_lp}
\end{figure*}

It is observed that the linear separability of CLIP's tokens is greater than that of those from pretrained VLMs' adapters on both coarse-grained and fine-grained classification datasets, as shown in Fig~\ref{fig:CLIP_lp}.

\begin{table*}
    \centering
    \vspace{-20pt}\tablestyle{3pt}{1.3}
    \caption{Details about dataset for both vision-language alignment and post-training.}
    \label{tbl:dataset}
    \begin{tabular}{ll}
         \hline
         vision-language alignment dataset for MobileVLM-v2\\
         \hline
         ShareGPT4V-PT(captioner)
         &
         1.2M

         \\
         (subset of following datasets:)
         \\
         \quad $\cdot$ LAION
         &

         \\
         \quad $\cdot$ CC\\
         \quad $\cdot$ SBU\\
         \quad $\cdot$ MS-COCO\\
         \hline
        post-training dataset (same for both)
        \\
        \hline
         LLaVA-Instruction
         &
         665K
         \\
         \quad $\cdot$ LLaVA\\
         \quad $\cdot$
         ShareGPT\\
         \quad $\cdot$ MS-COCO\\

         \quad $\cdot$ GQA\\

         \quad $\cdot$ OCR-VQA\\
         \quad
         $\cdot$
         VQAv2\\
         \quad
         $\cdot$
         OKVQA\\
         \quad
         $\cdot$
         A-OKVQA\\
         \quad
         $\cdot$
         TextCaps
         \\
         \quad
         $\cdot$
        RefCOCO
         \\
         \quad
         $\cdot$
         VG\\
         \bottomrule
    \end{tabular}
\end{table*}

\begin{table*}[h]
    \centering
    \tablestyle{7.9pt}{1.3}
    \caption{Training receipt.}
    \begin{tabular}{lll}
    \hline
    \multicolumn{3}{c}{vision-language alignment}
        \\
        \hline
         name
         &
         MobileVLM-v2 (1.7B/3B)
         \\
         vision-language alignment datasets
         &
          ShareGPT4V-PT
          \\
          trained parameters
          &
          Vision Adapter + LLM
          \\
          batchsize
          &
          256
          \\
          optimizer
          &
          AdamW
          \\
          learning rate
          &
          1$\times$10$^{-3}$
          \\
          scheduler
          &
          cosine
          \\
          vision tower
          &
          clip-vit-large-patch14-336
          \\
          LLM
          &MobileLLaMA-1.4B/2.7B-Chat
          \\
          adapter
          &
          ldpnetv2
          \\
          max sequence length
          &
          2048
          \\
          selected feature layer
          &
          second
          \\
          \hline \multicolumn{3}{c}{post-training}
          \\
          \hline
          post-training dataset
          &
          Instruction Tuning from LLaVA-v1.5
          \\
          batchsize
          &
          128
          \\
          optimizer
          &
          AdamW
          \\
          learning rate
          &
          4$\times$10$^{5}$
          \\
          scheduler
          &
          cosine
          \\
          default PEFT
          &
          LoRA
          \\
          \hline
          \multicolumn{3}{c}{
          related PEFT setting
          }
          \\
          \hline
          ranks (r) of LoRA
          &
          \multicolumn{2}{c}{32/64/128}
          \\
          $\alpha$ of LoRA
          &
          \multicolumn{2}{c}{
          256 (empirically fixed)
          }
          \\
          \bottomrule
    \end{tabular}
    \label{tbl_reciept}
\end{table*}

\section{Additional Ablation Experiments with Full Fine-tuning}
\begin{figure}[ht]
    \centering
    \vspace{-10pt}
    \includegraphics[width=0.99\linewidth]{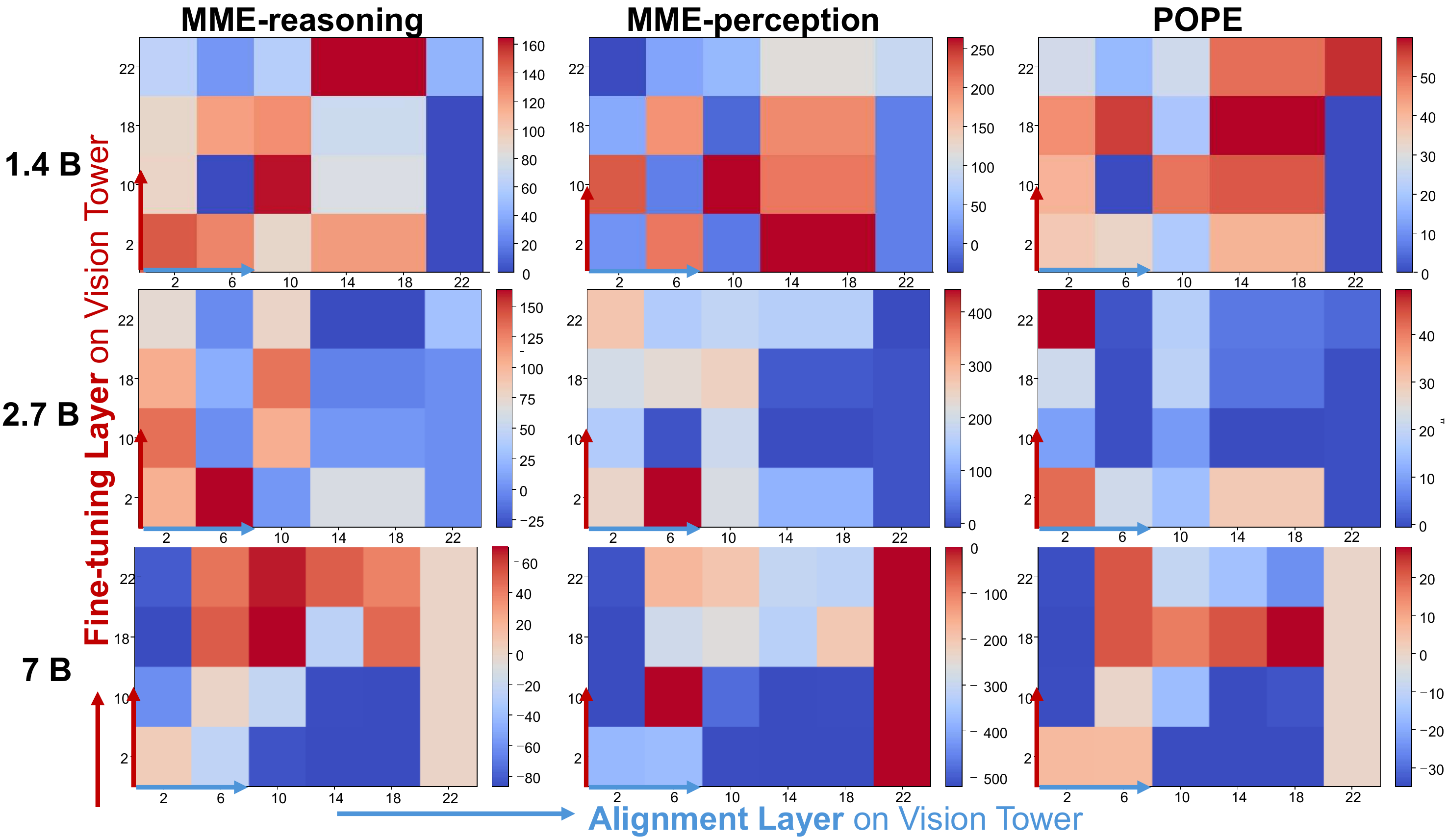}
    \vspace{-10pt}
    \caption{Depth preferences: the optimal alignment depth depends on the model scale, while patterns in a given model across similar benchmarks remain consistent.
    \emph{Horizontal axis}: the vision-tower layer used during the feature alignment stage, sampled from the original 23-layer CLIP encoder at layers \{2, 6, 10, 14, 18, 22\}.
    \emph{Vertical axis}: the vision-tower layer used during the instruction fine-tuning stage, selected from layers \{2, 10, 18, 22\} of the same 23-layer CLIP model. Default: 22 layer. MobileLLaMA-1.4B and -2.7B, Vicuna-7B are used as LLMs backbone.}
    \vspace{-20pt}
    \label{fig:abl-depth}
\end{figure}

\paragraph{Depth.}
In Figure~\ref{fig:abl-depth}, we conduct a grid search over alignment and instruction-tuning depths. Intuitively, deeper layers encode higher-level semantic representations, and aligning these may provide a clearer cross-modal bridge. However, the optimal alignment depth varies across model sizes, a phenomenon we term the \emph{alignment/fine-tuning depth preference}. Specifically, the depth yielding the best alignment is not necessarily optimal for fine-tuning. As illustrated in Figure~\ref{fig:abl-depth}, Vicuna-7B favors deeper layers for alignment and mid-level layers for fine-tuning on perception and hallucination benchmarks, whereas MobileLLaMA-1.4B and -2.7B both favor shallower-to-middle layers for alignment, with divergent fine-tuning depth preferences.

\paragraph{Learning scheduler.}
\begin{table}[ht]
    \centering
    \tablestyle{3.2pt}{1.2}
    \begin{tabular}{lrrrrrr}
    bench.
    &
    GQA
    &
    MME-p
    &
    POPE
    &
    TextVQA
    &
    MMB-dev
    &
    SQA
    \\
    \hline
         linear-full
         &
         61.15&1469.49&88.09&43.06&73.63&74.77
         \\
         linear-pevila
         &
         62.07&1545.05&87.36&46.72&73.02&74.15
         \\
         wsd-full
         &
         60.92&1482.24&87.42&42.34&73.54&74.37
         \\
         wsd-pevila
         &
         61.97&1482.24&87.09&47.34&75.69&75.41
    \end{tabular}
    \caption{Learning rates. Warmup-stable-decay: wsd.}
    \label{tab:rbt_lr}
\end{table}

Ablation about learning schedulers is shown in Tabel~\ref{tab:rbt_lr}. Consistent results are shown across different popular schedulers.

\section{Symbols}

Our employed symbols are summarized in Table~\ref{tab:symbol}
\begin{table}[t]
\caption{Symbols employed in our theoretical analysis.}
\label{tab:symbol}
\vspace{-10pt}
\tablestyle{30pt}{1.3}
    \begin{tabular}{ll}
\hline
Symbol & Meaning, Dimension and Shape \\ \hline
$i,j,k,m,n,s,t$ & Einstein sum. convention's indices \\
\multicolumn{2}{l}{\emph{~\textbf{Notice}: indices have no meaning but "duplicate$\rightarrow$sum".}}
\\
$O^i, H^j$ & Output and input (feature), range: $p$, $q$ \\
$O^{is}, H^{jt}$ & Kronecker Output/input, $p\times s$, $q\times t$ \\
$W_0^{\,i}{}_{j}, b^i$ & Base weight $p\times q$, bias $p$ \\
$A^{\,k}{}_{j}, B^{\,i}{}_{k}$ & Low-rank factors, $r\times q$, $p\times r$ \\
$B_1,B_2,A_1,A_2$ & Hadamard-case factors, $p\times r$, $r\times q$ \\
$C^{\,s}{}_{t}$ & Kronecker factor, $s\times t$ \\
$l^i(H;\theta)$ & Linear adaptation term, parameter $\theta$ \\
$F(O;\mathcal{D})$ & Scalar objective on dataset $\mathcal{D}$  \\
$\delta_i^j, \dot A$ & Kronecker delta, gradient flow of $A$ \\
$*, \otimes$ & Hadamard, Kronecker products \\
$M$ & Temporary variables for simplification \\
\hline
\end{tabular}
\vspace{-20pt}
\end{table}

\end{document}